\documentclass{article}
\usepackage{arxiv}

\usepackage[utf8]{inputenc}
\usepackage[T1]{fontenc}
\usepackage{microtype}
\usepackage{graphicx}
\usepackage{subcaption}
\usepackage{booktabs}
\usepackage{multirow}
\usepackage{hyperref}
\usepackage{url}
\usepackage{seqsplit}
\usepackage[numbers]{natbib}
\usepackage{amsmath}
\usepackage{amssymb}
\usepackage{mathtools}
\usepackage{amsthm}
\usepackage[capitalize,noabbrev]{cleveref}
\usepackage{algorithm}
\usepackage{algorithmic}

\title{OLEDLM: A Unified Language Model for OLED Molecular Design}
\author{
\normalfont
\parbox{0.95\textwidth}{%
\centering
\fontsize{11}{12}\selectfont
Fukang Wen\textsuperscript{1,5}, Yuchong Tang\textsuperscript{1,5}, Jingyuan Li\textsuperscript{2,3,5}, Beichen Wang\textsuperscript{1,5}, Yixuan Jiang\textsuperscript{1,5}, Xiaoyi Jiang\textsuperscript{1,5},\\
Yaxuan Liu\textsuperscript{4}, Shunyu Wang\textsuperscript{4}, Zuoqiang Shi\textsuperscript{1,2,5}, Yi Zhu\textsuperscript{1,2,5}, Yanan Zhu\textsuperscript{4,*,\ensuremath{\dagger}}, Pipi Hu\textsuperscript{2,5,*,\ensuremath{\ddagger}}\\[0.4em]
\textsuperscript{1}Tsinghua University\\
\textsuperscript{2}Beijing Institute of Mathematical Sciences and Applications\\
\textsuperscript{3}Wuhan University\\
\textsuperscript{4}Shenzhen MSU-BIT University \\
\textsuperscript{5}MathonAI Team
}
}

\begin{document}
\maketitle
\begingroup
\renewcommand{\thefootnote}{}
\footnotetext{\noindent\hspace{-1.8em}\textsuperscript{*}Corresponding author\\
\noindent\textsuperscript{\ensuremath{\dagger}}zhuyn@smbu.edu.cn \\
\noindent\textsuperscript{\ensuremath{\ddagger}}hpp@bimsa.cn
}
\endgroup

\begin{abstract}
The development of organic light-emitting diode (OLED) materials faces the compounded challenges of an astronomically large chemical space, stringent quantum-chemical constraints, and a scarcity of labeled data. Although the question of OLED generation is important, few models have been trained effectively for this specific domain. We propose an inverse molecular design framework based on causal language models: given target optoelectronic properties (e.g., excitation energy, oscillator strength), our model directly generates OLED SMILES sequences satisfying the specified constraints. We employ a multi-stage strategy: first, we establish a foundational chemical language model using a LLaMA-style transformer architecture. To the best of our knowledge, this represents the first successful adaptation of LLMs specifically for the OLED domain, bridging the gap between generic molecular generation and the stringent structural requirements of optoelectronic materials. Second, we fine-tune property predictors based on a BERT model pre-trained on our large-scale OLED dataset. Then, we perform Reinforcement Learning on our fine-tuned model, leveraging our property predictor, for better SMILES generation. Finally, through DFT verification, we demonstrate that our framework can efficiently navigate the OLED chemical space, generating novel candidates with high structural validity and optimized optoelectronic properties.
\end{abstract}

\section{Introduction}
\label{sec:intro}

The discovery of high-performance organic light-emitting diode (OLED) materials involves navigating an immensely vast chemical space to identify structures that simultaneously satisfy stringent optoelectronic constraints, such as specific singlet energy levels ($S_1$) and high oscillator strength ($f$). Traditional discovery pipelines, relying on high-throughput screening or heuristic modification, are often constrained by the prohibitive cost of Quantum Mechanics simulations and the limitations of human intuition. Recently, Generative AI has emerged as a transformative paradigm \citep{jumper2021alphafold, zeni2025mattergen}, shifting the focus from screening existing libraries to the de novo design of materials with targeted properties \citep{simonovsky2018graphvae, olivecrona2017molecular, hoogeboom2022edm, xu2022geodiff}.

In particular, Large Language Models trained on Simplified Molecular Input Line Entry System (SMILES) strings have demonstrated remarkable proficiency in learning the language of chemistry \citep{zhou2023unimol, guo2024augmented}. Recent studies suggest that Transformer-based architectures, originally designed for natural language, can effectively capture complex chemical syntax and structural semantics \citep{cui2024oled, kim2025deepblue}. In this work, we leverage a LLaMA-style Transformer for OLED molecule generation and employ Reinforcement Learning to align conditional generation, significantly improving the precision of property-targeted OLED molecule design.

Specifically, we adopt Group Relative Policy Optimization (GRPO) \citep{shao2024deepseekmath} to align the generative model with target properties. GRPO estimates advantages by sampling a group of molecular candidates for each condition and ranking them relative to the group mean, providing stable training without requiring a separate value network.

\textbf{Our contributions are as follows:}
\begin{enumerate}
    \item \textbf{OLED-Specific Chemical Language Model:} We curate a large-scale OLED-relevant molecular dataset and pre-train a LLaMA-style Transformer from scratch, establishing a foundational model capable of unconditional generation of valid, novel OLED molecular structures.
    \item \textbf{Domain-Adaptive Property Predictor:} We collect 10,000 OLED molecules with expensive DFT-computed optoelectronic properties ($S_1$, $f$) and train specialized BERT-based predictors that significantly outperform general-purpose encoders such as MoLFormer-XL on OLED-specific tasks.
    \item \textbf{Conditional Supervised Fine-Tuning:} We develop a property-conditioned SFT strategy using discrete control tokens, enabling the model to generate OLED molecules that satisfy user-specified $S_1$ and $f$ targets.
    \item \textbf{GRPO-Based Alignment:} Building upon our SFT model, we employ the property predictors as reward models within a Group Relative Policy Optimization (GRPO) framework, further improving the precision of conditional generation without the instability of critic-based RL methods.
\end{enumerate}

\section{Related Work}
\label{sec:related}

\subsection{Encoder-only Models for Molecular Representation}

The BERT architecture \citep{devlin2018bert}, originally designed for masked language modeling in NLP, has been successfully adapted to the chemical domain. Models such as ChemBERTa \citep{chithrananda2020chemberta} and MoLFormer-XL \citep{ross2022large} utilize bidirectional attention mechanisms to learn rich molecular representations from massive SMILES datasets. These models typically serve as powerful encoders for downstream predictive tasks. In our framework, distinct from these general-purpose encoders, we leverage a customized BERT-style architecture specifically as a property predictor. By training on domain-specific OLED datasets, our BERT module acts as a rigorous evaluator, providing high-fidelity feedback to guide the generation process.

\subsection{Large Language Models for De Novo Design}

Autoregressive Transformer models, exemplified by LLaMA \citep{touvron2023llama}, have demonstrated exceptional capability in modeling discrete sequential data. In the context of material science, SMILES strings can be treated as a ``chemical language.'' Recent studies have shown that decoder-only Transformers can learn the syntax of valid molecules and generate novel structures \citep{flam2022language}. Our work adopts the LLaMA architecture to ensure training stability and long-range dependency modeling in complex molecular sequences.

\subsection{Conditional Generation and Supervised Fine-Tuning}

The ``Pretrain-Finetune'' paradigm has become the de facto standard in generative AI. While ``Instruction Tuning'' \citep{wei2021finetuned} enables models to follow natural language commands, molecular generation often requires precise control over physical properties rather than open-ended dialogue. Approaches like CTRL \citep{keskar2019ctrl} or property-conditioned fine-tuning introduce control codes into the input sequence. In this work, we implement a Conditional Supervised Fine-Tuning (SFT) strategy. Instead of vague natural language prompts, we condition the LLaMA model on quantized property tokens (e.g., specific $S_1$ and $f$ ranges), effectively teaching the model to map desired optoelectronic criteria to specific chemical structures.

\subsection{Reinforcement Learning for Alignment}

A fundamental limitation of SFT for property-conditioned generation is the sparsity of supervision signal. The labeled dataset of (property, SMILES) pairs is inherently small compared to the vast chemical space. If trained for too few epochs, the model fails to learn the implicit mapping between property tokens and molecular structures, resulting in poor conditional generation. Conversely, training for too many epochs leads to memorization and mode collapse, where the model simply reproduces training examples rather than generalizing. To address this dilemma, we adopt Reinforcement Learning (RL) after SFT. Instead of directly exposing the model to (property, SMILES) pairs, RL uses a property predictor as a reward signal to guide generation, providing dense, differentiable feedback without overfitting to specific examples. Traditional RL methods like PPO \citep{schulman2017proximal} rely on a Value Network (Critic) to estimate advantages, which is difficult to train in the rugged landscape of chemical properties. Recently, Group Relative Policy Optimization, introduced in DeepSeekMath \citep{shao2024deepseekmath}, proposed a critic-free approach that estimates advantages by sampling a group of outputs and normalizing rewards relative to the group mean. We identify GRPO as uniquely suited for molecular discovery, as it effectively solves the ``Critic Bottleneck'' by leveraging the generative model itself to provide a stable, group-relative baseline.

\subsection{AI for OLED Materials Discovery}

Machine learning has been increasingly applied to OLED discovery, primarily for screening existing libraries or optimizing continuous representations (e.g., VAE latent spaces) \citep{gomez2018automatic}. However, these methods often struggle with validity decoding or lack precise control over specific optoelectronic properties. De novo generative approaches targeting strict multi-objective optimization---specifically high oscillator strength ($f$) for efficiency and precise singlet energy ($S_1$) for color purity---remain scarce. To the best of our knowledge, our work is the first to integrate GRPO-based Reinforcement Learning with a LLaMA-style backbone to achieve high-fidelity de novo design of OLED emitters that satisfy these conflicting constraints.

\section{Methodology}
\label{sec:method}

\begin{figure}[t]
\centering
\includegraphics[width=0.8\textwidth]{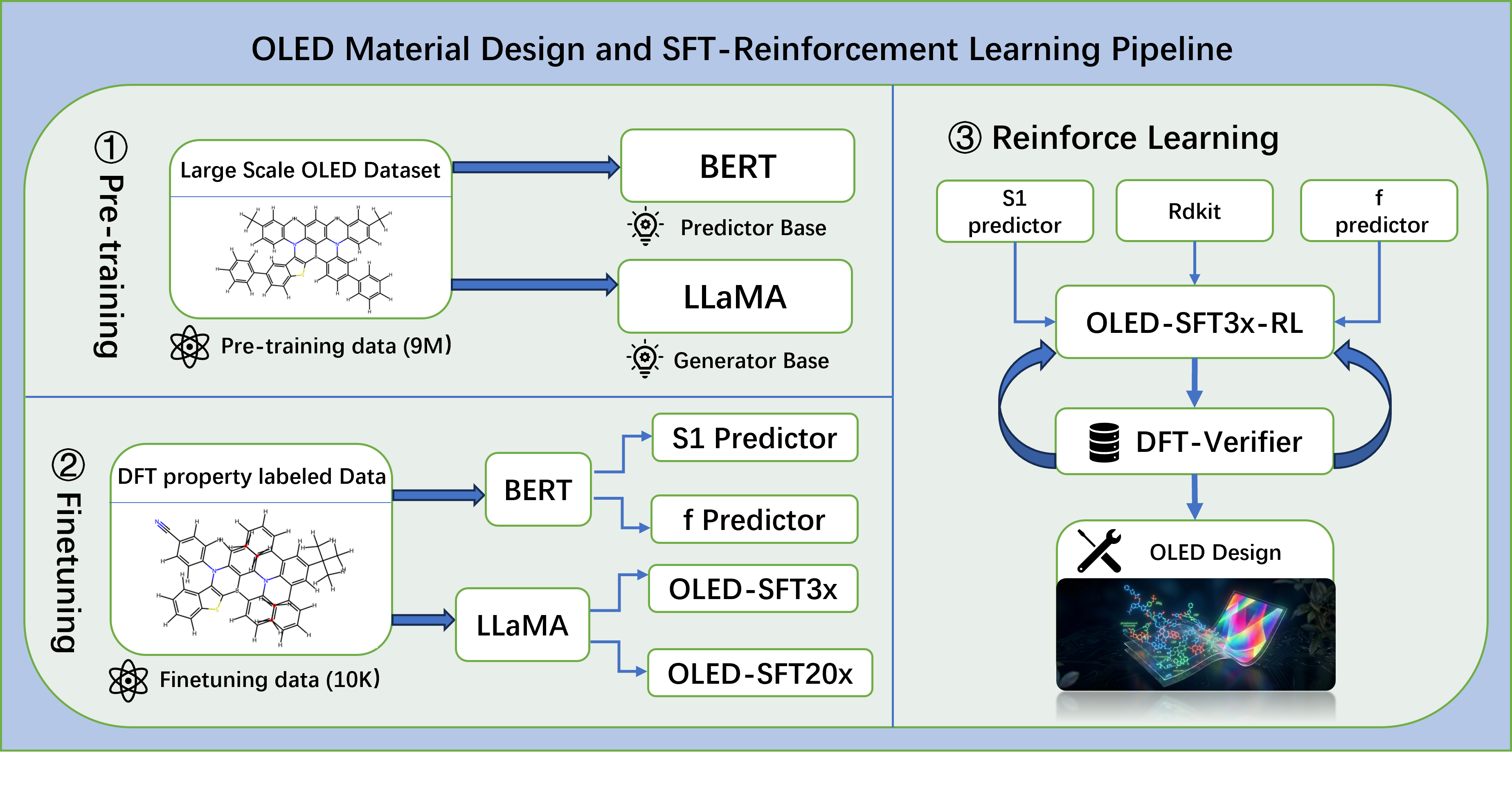}
\caption{Overview of our unified bidirectional OLED molecular design framework. Starting from a 9M SMILES pre-training corpus, we train two complementary models via self-supervised learning: (1) OLED-BERT for property prediction, and (2) OLED-LLaMA for molecular generation. The 9k DFT-verified dataset enables fine-tuning OLED-BERT into specialized $S_1$ and $f$ predictors, while guiding conditional SFT of OLED-LLaMA. The predictors then serve as reward models to further refine the generator via GRPO, achieving precise property-conditioned molecular design.}
\label{fig:overview}
\end{figure}

\subsection{Domain-Adaptive Property Predictor}

To guide the generative model toward high-performance OLED candidates, we require a robust reward model capable of accurately predicting singlet energy ($S_1$) and oscillator strength ($f$). While general-purpose chemical language models like MoLFormer-XL exist, they are typically pre-trained on broad, heterogeneous datasets (e.g., PubChem, ZINC) that may not adequately capture the specific electronic structure nuances of conjugated OLED materials. Our experiments demonstrated that fine-tuning such off-the-shelf models on OLED-specific data yielded significantly higher prediction errors compared to our domain-adaptive approach (see Section~\ref{sec:experiments}).

Consequently, we adopted a domain-adaptive pre-training strategy. We trained a BERT-style encoder from scratch on a curated dataset of 9 million SMILES strings using the Masked Language Modeling (MLM) objective. This allows the model to learn a specialized chemical vocabulary and syntax relevant to organic semiconductors. For the downstream property prediction task, we employed a partial fine-tuning approach: we froze the initial layers of the encoder and fine-tuned only the final transformer layers and the regression head on our labeled dataset of 9,000 DFT-calculated samples. This strategy effectively mitigates overfitting to the small amount of labeled data while retaining the rich feature extraction capabilities learned during pre-training. Our internal evaluations show this specialized predictor achieves a Mean Absolute Error (MAE) of 0.12 eV for $S_1$ ($R^2=0.69$) and 0.06 for $f$ ($R^2=0.68$), providing reliable feedback for the subsequent reinforcement learning stage.

\subsection{Generative Backbone and Tokenization}

We model molecular generation as a causal language modeling task. Our backbone is a LLaMA-style Transformer, incorporating Rotary Positional Embeddings, RMSNorm, and SwiGLU activation functions. These architectural choices ensure training stability and effective long-range dependency modeling, which are critical for generating valid, complex molecular strings. We utilize a custom SMILES tokenizer designed to treat chemical substructures and symbols as discrete tokens, enabling the model to process molecular strings as a structured language. Similar to its success in natural language, this architecture demonstrates strong generalization capabilities in learning the probabilistic distribution of valid chemical structures.

\subsection{Conditional Supervised Fine-Tuning (SFT)}

To align the model's output with desired physical properties, we implement Conditional Supervised Fine-Tuning. Unlike standard ``instruction tuning'' which typically relies on natural language prompts, our approach explicitly conditions the generation on quantized property values using discrete control codes. We map continuous properties ($S_1$, $f$) into specific tokens (e.g., \texttt{<s1>}, \texttt{<num\_2>}, \texttt{<dot>}, \texttt{<res\_50>}) which are prepended to the SMILES sequence.

We construct a supervised dataset where each OLED molecule is paired with its ground-truth property tokens. By optimizing the conditional log-likelihood $\log P(\text{SMILES} | \text{Properties})$, the model learns to associate specific property constraints with corresponding structural motifs. This stage provides a ``warm-start'' policy that roughly adheres to target conditions, serving as a stable initialization for reinforcement learning.

\paragraph{Mitigating Memorization in SFT.}
In practice, we encountered a critical challenge: when training exclusively on the small amount of property-labeled dataset, the model rapidly memorizes specific SMILES strings rather than learning the underlying structure-property relationships. This memorization manifests as severely degraded generation diversity, rendering the model unsuitable for subsequent RL fine-tuning. To address this, we employ a \textit{mixed-data training} strategy: we augment the SFT dataset with 10$\times$ the volume of unlabeled SMILES from the pre-training corpus. Crucially, we \textit{re-sample} this auxiliary data at each epoch, ensuring the model encounters diverse molecular structures throughout training. This approach effectively prevents catastrophic forgetting of the learned chemical grammar while mitigating overfitting to the small labeled set. Empirically, we observe that direct SFT on high-quality DFT-labeled data leads to rapid overfitting within a few epochs, whereas our mixed-data strategy maintains the next-token prediction accuracy at levels comparable to the pre-trained model, indicating preserved generalization capability.

\subsection{Alignment via Group Relative Policy Optimization}
\label{sec:grpo}

A fundamental limitation of SFT for property-conditioned molecular generation is the training dilemma. The LLaMA architecture excels at memorizing specific SMILES strings due to its powerful autoregressive capability. However, this strength becomes a weakness during conditional fine-tuning: when trained with property tokens prepended to SMILES sequences, the model tends to memorize the exact (property, SMILES) pairs from training data rather than learning the underlying structure-property relationships. Training for too few epochs results in poor conditional generation, while training for too many epochs causes mode collapse. More critically, introducing property tokens into the training sequences disrupts the learned chemical syntax, leading to a significant drop in SMILES validity. To address this dilemma, we adopt Reinforcement Learning, which provides dense feedback through a reward model without requiring the model to memorize specific property-SMILES pairs.

\paragraph{GRPO: Group Relative Policy Optimization.}
We employ Group Relative Policy Optimization (GRPO). Instead of estimating the value of incomplete states, GRPO samples a group of \textit{complete} outputs $\{o_1, \ldots, o_G\}$ for each condition $q$ and uses their relative performance as a self-referential baseline. The advantage for each sample is:
\begin{equation}
A_i = \frac{R(o_i) - \text{mean}(\{R(o_j)\}_{j=1}^G)}{\text{std}(\{R(o_j)\}_{j=1}^G) + \epsilon}
\label{eq:advantage}
\end{equation}
where $\epsilon$ is a small constant for numerical stability. By comparing complete OLED molecules against each other, GRPO sidesteps the unstable value estimation step entirely.

\paragraph{Reward Function.}
Our reward function $R(x)$ for a generated OLED molecule $x$ with target properties $(S_1^*, f^*)$ is defined as:
\begin{equation}
\begin{split}
R(x) = {}& \mathbb{I}_{\text{valid}}(x) \cdot \big[ -\alpha |S_1(x) - S_1^*| - \gamma |f(x) - f^*| \big] 
 + (1 - \mathbb{I}_{\text{valid}}(x)) \cdot \lambda_{\text{penalty}}
\end{split}
\label{eq:reward}
\end{equation}
where $\mathbb{I}_{\text{valid}}(x)$ is an indicator function for RDKit-parseable SMILES, $S_1(x)$ and $f(x)$ are the BERT predictor estimates, and $\lambda_{\text{penalty}} < 0$ penalizes invalid structures.

\paragraph{KL Regularization.}
A common failure mode in molecular RL is mode collapse, where the policy converges to a small set of high-reward OLED molecules. To mitigate this, we introduce KL divergence regularization against the reference SFT model $\pi_{\text{ref}}$, ensuring the agent explores the chemical space without exploiting ``holes'' in the reward model.

The full GRPO objective is:
\begin{equation}
\begin{split}
\mathcal{L}_{\text{GRPO}}(\theta) = \mathbb{E}_{q, \{o_i\}} \Bigg[ \frac{1}{G} \sum_{i=1}^{G} \bigg( & \min \big( \rho_i A_i, \text{clip}(\rho_i) A_i \big) 
 - \beta D_{\text{KL}}(\pi_\theta \| \pi_{\text{ref}}) \bigg) \Bigg]
\end{split}
\label{eq:grpo}
\end{equation}
where $\rho_i(\theta) = \frac{\pi_\theta(o_i|q)}{\pi_{\theta_{old}}(o_i|q)}$ is the importance sampling ratio and $\beta$ controls the regularization strength. In our implementation, we approximate the KL term using the absolute difference in log-probabilities for training stability.

\section{Experiments}
\label{sec:experiments}

\subsection{Experimental Setup}

\subsubsection{Dataset}
To construct a dedicated dataset for model pretraining and fine-tuning, we independently generated an OLED synthesis dataset with a scale of 9 million molecules. All molecular structures in this dataset were constructed in accordance with chemical rules, following a specific generation protocol: novel molecular structures were synthesized via site-specific substitution based on common molecular parent cores and substituents. In addition, we acquired a 10,000-scale OLED dataset by extracting relevant information from patents and literature, for which we obtained Singlet Energy ($S_1$) and Oscillator Strength ($f$) as dataset labels through density functional theory (DFT) calculations under the conditions of the B3LYP functional and the 6-31G** basis set. This labeled dataset was split into 9,000 samples for training and 1,000 samples for evaluation (500 for validation, 500 for testing).

All data are presented in the form of canonical SMILES. As a standardized and unique version of the Simplified Molecular-Input Line-Entry System (SMILES), canonical SMILES generates a unique string for the same molecule via a fixed algorithm, thus eliminating the ambiguity of multiple encodings for identical molecular structures. This ensures the uniqueness and uniformity of molecular structure characterization, effectively avoiding errors in data statistics and structure matching.

\subsubsection{Model Architecture}
Our generative backbone is a LLaMA-style Transformer with the following specifications: hidden dimension of 768, 12 transformer layers, 12 attention heads, and SwiGLU activation in the feed-forward network. We employ Rotary Position Embeddings (RoPE) with $\theta=10000$ and RMSNorm for layer normalization. The model supports sequences up to 512 tokens and contains approximately 100M parameters. The tokenizer uses a custom SMILES vocabulary of 147 base tokens extended to 265 tokens with property control codes (e.g., \texttt{<s1>}, \texttt{<f>}, \texttt{<num\_X>}, \texttt{<dot>}, \texttt{<res\_YY>}) for conditional generation.

For the domain-adaptive property predictor, we train a BERT-style encoder (768 hidden dimensions, 12 layers, 12 heads, $\sim$86M parameters) using Masked Language Modeling on the pre-training corpus. The predictor is then fine-tuned for $S_1$ and $f$ regression tasks separately, with mean pooling over token embeddings followed by a 2-layer MLP regression head.

\subsubsection{Baselines}
We compare four generation settings. The \textit{Unconditional} model samples from the pretrained generator without property controls. \textit{SFT} is conditionally fine-tuned for 3 epochs, while \textit{SFT-20x} is trained for 20 epochs to expose the effect of extended supervised training. \textit{GRPO} starts from the SFT policy and is optimized with the proposed reward. PPO and an LSTM generator adapted from LumiGen are evaluated separately as controlled algorithmic and architectural comparisons, respectively.

\subsubsection{GRPO Configuration}
Our GRPO implementation employs a group size of $G=20$ samples per property condition. We sample property targets from the empirical distribution of the training set to ensure the model learns practically relevant OLED profiles. We set the KL regularization coefficient $\beta=0.5$ and impose a validity soft-penalty of $-1.0$ (scaled) for SMILES that fail RDKit parsing. Training proceeds for 1,000 steps with a learning rate of $10^{-5}$ using the AdamW optimizer.

\subsubsection{Evaluation Metrics}
We evaluate all methods on:
\begin{itemize}
    \item \textbf{Validity (\%):} Percentage of generated SMILES that pass RDKit parsing and represent valid molecular structures.
    \item \textbf{$S_1$ MAE (eV):} Mean absolute error between the BERT predictor's estimate of $S_1$ on the generated OLED molecule and the target value.
    \item \textbf{$f$ MAE:} Mean absolute error for oscillator strength prediction.
    \item \textbf{Uniqueness (\%):} Percentage of unique valid OLED molecules among all valid generations.
    \item \textbf{Novelty (\%):} Percentage of valid OLED molecules not present in the training set.
\end{itemize}

\subsubsection{Evaluation Protocol}
We evaluate the main generation benchmark by sampling 10,000 SMILES sequences per method. For the unconditional model, we sample directly from the pretrained LLaMA backbone starting from the \texttt{<bos>} token. For the conditional models, we sample target property pairs $(S_1, f)$ from the empirical distribution of the training set, encode each pair as property tokens, and generate the corresponding SMILES. All generations use temperature $T=1.0$, top-$k=50$, and top-$p=0.95$ sampling. We compute validity via RDKit parsing, property alignment via BERT predictor inference, uniqueness among valid generations, and novelty against the labeled training set. The PPO--GRPO comparison uses 100 matched generations and is therefore treated as an algorithmic ablation rather than a replacement for the 10,000-sample benchmark.

\subsection{Main Results}

\begin{table}[H]
\caption{Comparison of generation quality across methods. GRPO achieves the best validity while maintaining high novelty. SFT-20x shows strong property alignment but suffers from memorization (low novelty). Results averaged over 10,000 generated samples.}
\label{tab:results}
\centering
\small
\begin{tabular}{@{}lccccc@{}}
\toprule
Method & Valid. & $S_1$ MAE & $f$ MAE & Uniq. & Novel \\
\midrule
Unconditional & 74.9 & N/A & N/A & 100.0 & 100.0 \\
SFT           & 82.9 & 0.28 & 0.20 & 90.2 & 89.4 \\
SFT-20x       & 91.9 & \textbf{0.19} & \textbf{0.12} & 75.8 & 58.0 \\
\textbf{GRPO} & \textbf{98.3} & 0.25 & 0.15 & 67.2 & \textbf{94.4} \\
\bottomrule
\end{tabular}
\end{table}

\paragraph{GRPO vs. SFT.} The SFT baseline (3 epochs) achieves moderate validity (82.9\%) with property alignment errors of $S_1$ MAE: 0.28 eV and $f$ MAE: 0.20. GRPO dramatically improves validity to 98.3\% while reducing $f$ MAE by 25\% (0.20 $\to$ 0.15). This demonstrates that RL-based alignment with explicit validity rewards can substantially enhance structural correctness beyond supervised learning.

\paragraph{The SFT Memorization Problem.} Extended SFT training (SFT-20x, 20 epochs) reveals a critical limitation: while it achieves the best property alignment ($S_1$ MAE: 0.19, $f$ MAE: 0.12), the novelty rate collapses to 58.0\%, indicating severe memorization of training examples. This confirms our hypothesis that SFT struggles to learn generalizable structure-property relationships, instead memorizing specific (property, SMILES) pairs from the limited labeled dataset.

\paragraph{Property--novelty trade-off.} The results expose different failure modes for extended SFT and RL. SFT-20x achieves the lowest predictor errors ($S_1$ MAE: 0.19, $f$ MAE: 0.12), but its novelty falls from 89.4\% for SFT to 58.0\%, consistent with increased retrieval of labeled training molecules. GRPO has lower uniqueness than SFT (67.2\% versus 90.2\%), indicating concentration on high-reward structural motifs, but retains 94.4\% novelty. Thus, the lower uniqueness of GRPO should not by itself be interpreted as training-set memorization. Together with its 98.3\% validity and competitive property alignment, this result suggests that GRPO offers a more favorable trade-off for de novo discovery, where generating novel candidates is an essential objective.

\subsection{DFT Validation}
\label{sec:dft_validation}

\begin{figure}[h]
\centering
\includegraphics[width=0.85\textwidth]{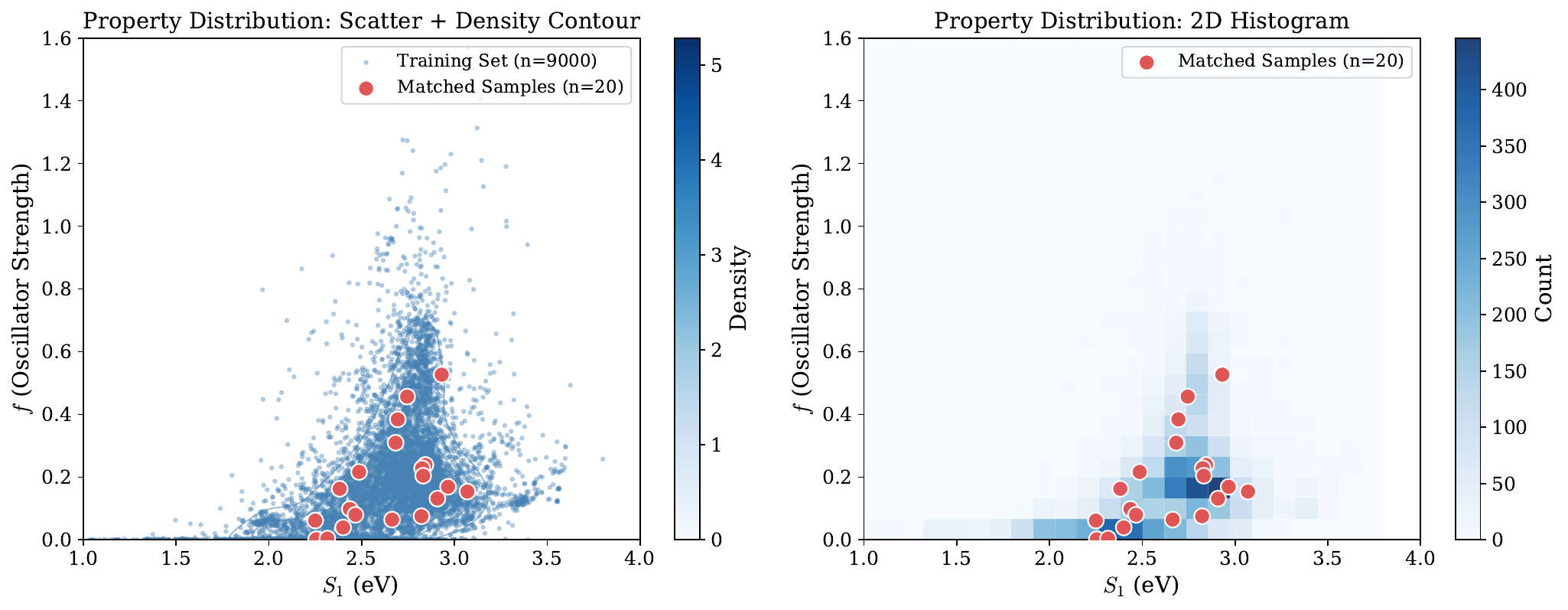}
\caption{Joint property distribution for DFT validation target selection. Left: Scatter plot with kernel density estimation contours showing the training set distribution (blue, n=9,000) and matched DFT validation samples (red, n=20). Right: 2D histogram with matched sample overlay. The 20 target pairs span diverse regions of the property space, covering $S_1 \in [2.25, 3.07]$ eV and $f \in [0.002, 0.53]$.}
\label{fig:dft_dist}
\end{figure}

To provide rigorous experimental validation beyond predictor-based evaluation, we conduct DFT calculations on generated OLED molecules. We sample target property pairs $(S_1^*, f^*)$ that span the training distribution, covering diverse regions of the property space (Figure~\ref{fig:dft_dist}).

Using identical target conditions, we generate OLED molecules from three models: (1) the GRPO-aligned model, (2) the SFT model trained for 3 epochs (SFT), and (3) the SFT model trained for 20 epochs (SFT-20x). Due to the prohibitive computational cost of DFT calculations (each molecule requires geometry optimization and TD-DFT excited state calculations at the B3LYP/6-31G** level), the initial submission reported only four matched converged conditions. We subsequently completed the remaining calculations and now report 20 converged molecules per method in Table~\ref{tab:dft_validation}.

\begin{table}[h]
\caption{Expanded DFT validation results comparing ground-truth quantum-chemical properties against target values. We report the mean absolute error and variance over 20 converged molecules per method.}
\label{tab:dft_validation}
\vskip 0.15in
\begin{center}
\begin{small}
\begin{sc}
\begin{tabular}{@{}lccccc@{}}
\toprule
Method & $n$ & $S_1$ MAE (eV) & $f$ MAE & $S_1$ Var & $f$ Var \\
\midrule
SFT-20x & 20 & \textbf{0.138} & \textbf{0.076} & \textbf{0.045} & \textbf{0.010} \\
GRPO    & 20 & 0.260 & 0.134 & 0.095 & 0.026 \\
SFT     & 20 & 0.339 & 0.166 & 0.181 & 0.059 \\
\bottomrule
\end{tabular}
\end{sc}
\end{small}
\end{center}
\vskip -0.1in
\end{table}

The expanded validation resolves the original sample-size limitation: 20 molecules per method reached convergence and were evaluated independently of the BERT reward model. SFT-20x achieves the lowest DFT error, but this result should be interpreted alongside its low novelty (58.0\%), consistent with memorization of training examples. GRPO maintains 94.4\% novelty while achieving lower DFT error than SFT. Thus, the expanded DFT results support the claim that the predictor-based reward provides useful signal on generated molecules, while also showing that SFT-20x remains stronger when only property accuracy is considered.

\subsection{Baseline and Algorithm Comparison}
\label{sec:additional_results}

\subsubsection{GRPO versus PPO}
To isolate the contribution of the policy-optimization algorithm, we compare GRPO and PPO using the same LLaMA+SFT checkpoint, reward model, KL coefficient ($\beta=0.5$), and 1,000 training steps. On a matched set of 100 generated molecules, both methods achieve 100\% validity. GRPO obtains lower property errors, whereas PPO produces higher uniqueness and novelty.

The controlled comparison is reported in Table~\ref{tab:ppo_comparison}. Because it uses 100 rather than 10,000 generations, it should be interpreted as an algorithmic comparison rather than as the main generation benchmark.

\begin{table}[h]
\caption{Controlled comparison between GRPO and PPO on 100 generated molecules. The two methods use the same generator initialization, reward model, and training configuration. GRPO provides lower property errors, whereas PPO retains higher uniqueness and novelty.}
\label{tab:ppo_comparison}
\centering
\small
\begin{tabular}{@{}lcc@{}}
\toprule
Metric & GRPO & PPO \\
\midrule
Validity (\%) & \textbf{100} & \textbf{100} \\
$S_1$ MAE (eV) & \textbf{0.250} & 0.267 \\
$f$ MAE & \textbf{0.130} & 0.178 \\
Uniqueness (\%) & 89 & \textbf{99} \\
Novelty (\%) & 86 & \textbf{93} \\
\bottomrule
\end{tabular}
\end{table}

GRPO therefore provides better target-property control in this matched comparison, while PPO retains a diversity advantage. PPO also required explicit implementation checks for per-token importance ratios, KL computation, and dropout consistency to remain stable, whereas the GRPO training configuration was stable without these additional interventions.

\subsubsection{LSTM/LumiGen observation}
For a broader architecture comparison, we also tested an LSTM-based conditional generator adapted from LumiGen on the OLED data. The LSTM collapsed on the substantially longer OLED SMILES sequences (approximately 80 characters), producing only nine trivial fragments such as ``CC''. Consequently, no meaningful generation metrics could be computed. We treat this result as an architecture feasibility observation rather than a quantitative baseline.

\subsection{Ablation Studies}
\label{sec:predictor_ablation}

\subsubsection{Reward Model / Predictor Ablation}
The quality of the BERT-based property predictor directly impacts GRPO's effectiveness. We compare our domain-adaptive BERT against general-purpose molecular encoders, varying the number of unfrozen layers, the regression-head dimension, and the pre-training mask ratio. The results are reported on the held-out validation set of 500 OLED molecules.

\begin{table}[t]
\caption{Comprehensive predictor ablation results on the held-out validation set (500 OLED molecules). We compare four backbone models across configurations: unfrozen layers (1 or 2) and MLP hidden dimensions. Our domain-adaptive OLED-BERT models significantly outperform general-purpose encoders. Bold indicates best; underline indicates second best.}
\label{tab:predictor}
\vskip 0.15in
\begin{center}
\begin{small}
\begin{sc}
\begin{tabular}{@{}llcccccc@{}}
\toprule
Model & Pre-training Data & Unfreeze & Hidden & $S_1$ MAE & $S_1$ $R^2$ & $f$ MAE & $f$ $R^2$ \\
\midrule
\multirow{6}{*}{ChemBERTa} & \multirow{6}{*}{ZINC (77M)}
  & 1 & 256 & 0.202 & 0.34 & 0.108 & 0.30 \\
& & 1 & 512 & 0.193 & 0.38 & 0.108 & 0.32 \\
& & 1 & 768 & 0.192 & 0.40 & 0.108 & 0.35 \\
& & 2 & 256 & 0.202 & 0.34 & 0.108 & 0.30 \\
& & 2 & 512 & 0.193 & 0.38 & 0.108 & 0.32 \\
& & 2 & 768 & 0.192 & 0.40 & 0.108 & 0.35 \\
\midrule
\multirow{6}{*}{MoLFormer-XL} & \multirow{6}{*}{PubChem (1.1B)}
  & 1 & 256 & 0.153 & 0.60 & 0.081 & 0.55 \\
& & 1 & 512 & 0.145 & 0.62 & 0.083 & 0.53 \\
& & 1 & 768 & 0.144 & 0.63 & 0.080 & 0.56 \\
& & 2 & 256 & 0.139 & 0.65 & \underline{0.075} & \underline{0.61} \\
& & 2 & 512 & 0.137 & 0.65 & 0.082 & 0.55 \\
& & 2 & 768 & \underline{0.134} & \underline{0.65} & 0.081 & 0.56 \\
\midrule
\multirow{6}{*}{OLED-BERT-0.15} & \multirow{6}{*}{OLED (9M)}
  & 1 & 256 & 0.148 & 0.60 & 0.071 & 0.63 \\
& & 1 & 512 & 0.124 & 0.69 & 0.071 & 0.64 \\
& & 1 & 768 & \textbf{0.124} & 0.69 & 0.071 & 0.64 \\
& & 2 & 256 & 0.135 & 0.64 & 0.091 & 0.52 \\
& & 2 & 512 & 0.190 & 0.40 & 0.072 & 0.64 \\
& & 2 & 768 & 0.260 & 0.26 & \textbf{0.067} & \textbf{0.68} \\
\midrule
\multirow{6}{*}{OLED-BERT-0.25} & \multirow{6}{*}{OLED (9M)}
  & 1 & 256 & 0.148 & 0.60 & 0.082 & 0.58 \\
& & 1 & 512 & 0.129 & 0.67 & 0.073 & 0.64 \\
& & 1 & 768 & \textbf{0.124} & \textbf{0.69} & 0.073 & 0.64 \\
& & 2 & 256 & 0.137 & 0.63 & 0.067 & 0.68 \\
& & 2 & 512 & 0.135 & 0.64 & 0.065 & 0.68 \\
& & 2 & 768 & 0.131 & 0.66 & \underline{0.064} & \underline{0.68} \\
\bottomrule
\end{tabular}
\end{sc}
\end{small}
\end{center}
\vskip -0.1in
\end{table}

\paragraph{Domain-adaptive pretraining.} Table~\ref{tab:predictor} reveals a performance gap between general-purpose molecular encoders and our domain-adaptive models. ChemBERTa, pre-trained on 77M drug-like molecules from ZINC, achieves only $R^2 \approx 0.34$--0.40 for $S_1$ and $R^2 \approx 0.30$--0.35 for $f$, while MoLFormer-XL trained on 1.1B molecules from PubChem and ZINC reaches $R^2 \approx 0.65$ for $S_1$ and $R^2 \approx 0.61$ for $f$. In contrast, our OLED-BERT models achieve $R^2 \approx 0.69$ for $S_1$ and $R^2 \approx 0.68$ for $f$, despite being pre-trained on only 9M OLED-relevant SMILES.

This performance advantage is consistent with distribution shift between general molecular databases and OLED materials. OLED emitters occupy a specialized chemical subspace characterized by extended $\pi$-conjugated systems, boron--nitrogen heterocycles, and donor--acceptor motifs that may be underrepresented in drug-focused datasets. Domain-adaptive pretraining can therefore provide a more relevant reward signal for downstream RL.

\paragraph{Architecture and held-out test results.} The ablation results suggest that larger hidden dimensions improve performance, while the optimal number of unfrozen layers varies by property: $S_1$ prediction benefits from minimal fine-tuning, whereas $f$ prediction improves with more adaptation. OLED-BERT-0.25 slightly outperforms OLED-BERT-0.15 for oscillator-strength prediction, suggesting that a higher masking ratio may encourage learning of global molecular features.

Because the predictor is used as a reward model, we separately evaluate the selected predictor on a held-out test set of 500 molecules with DFT ground truth. It achieves an $S_1$ MAE of 0.158~eV with $R^2=0.51$ and an $f$ MAE of 0.078 with $R^2=0.58$. These results do not eliminate distribution shift, but they show that the reward model provides a non-trivial proxy before the independent DFT evaluation.

\subsubsection{Reward and Training Objective Ablations}

\paragraph{Validity Reward Contribution.} We ablate the discrete validity penalty term in the reward function.

\begin{table}[h]
\caption{Impact of validity penalty on GRPO performance. Explicit validity rewards improve structural correctness without harming property alignment.}
\label{tab:validity_ablation}
\vskip 0.15in
\begin{center}
\begin{small}
\begin{sc}
\begin{tabular}{@{}lcccc@{}}
\toprule
Validity Reward & Valid. & $S_1$ MAE & $f$ MAE & Uniq. \\
\midrule
With penalty    & \textbf{97.4} & 0.29 & 0.20 & 93.2 \\
Without penalty & 88.5 & 0.29 & 0.20 & 96.9 \\
\bottomrule
\end{tabular}
\end{sc}
\end{small}
\end{center}
\vskip -0.1in
\end{table}

Without validity penalties, the model occasionally generates syntactically incorrect SMILES that happen to receive favorable property predictions from the reward model, achieving only 88.5\% validity. Adding an explicit validity penalty significantly improves structural correctness to 97.4\% while maintaining identical property alignment ($S_1$ MAE: 0.29, $f$ MAE: 0.20). Interestingly, removing the validity constraint slightly increases uniqueness (96.9\% vs 93.2\%), suggesting that some diverse structural patterns may be discouraged by strict validity enforcement.

\paragraph{Effect of KL Regularization Coefficient $\beta$.} The KL penalty balances property optimization against distribution drift from the SFT reference policy. Table~\ref{tab:kl_ablation} examines this trade-off.

\begin{table}[h]
\caption{Effect of KL regularization strength $\beta$ on GRPO performance. Moderate regularization ($\beta=0.5$) achieves optimal balance.}
\label{tab:kl_ablation}
\vskip 0.15in
\begin{center}
\begin{small}
\begin{sc}
\begin{tabular}{@{}lcccc@{}}
\toprule
$\beta$ & Valid. & $S_1$ MAE & $f$ MAE & Uniq. \\
\midrule
0.0 (No KL) & 98.8 & 0.28 & 0.18 & 84.2 \\
0.1         & 98.4 & 0.29 & 0.20 & 88.9 \\
0.5         & 97.5 & \textbf{0.27} & 0.21 & 93.8 \\
1.0         & 96.4 & 0.29 & 0.21 & 95.7 \\
\bottomrule
\end{tabular}
\end{sc}
\end{small}
\end{center}
\vskip -0.1in
\end{table}

Without KL regularization ($\beta=0$), the model achieves the lowest $f$ MAE (0.18) and highest validity (98.8\%), but shows reduced uniqueness (84.2\%) compared to higher $\beta$ values. Increasing $\beta$ progressively improves uniqueness: from 84.2\% at $\beta=0$ to 95.7\% at $\beta=1.0$. Moderate regularization ($\beta=0.5$) achieves the best $S_1$ MAE (0.27) while maintaining good diversity (93.8\% uniqueness). Higher values ($\beta=1.0$) slightly degrade validity (96.4\%) and property alignment, as the policy remains too constrained to the SFT distribution.

\subsection{Molecular Quality and Diversity Analysis}
\label{sec:molecular_quality}

We additionally evaluate synthetic accessibility and distributional diversity on 150 DFT-validated molecules, with 50 molecules from each of GRPO, SFT, and SFT-20x. These analyses are complementary to the property-alignment metrics: SA scores provide a computational proxy for synthetic accessibility, while scaffold- and fragment-level metrics characterize structural novelty and reuse of chemically plausible building blocks.

\begin{table}[h]
\caption{Synthetic accessibility (SA) score statistics on generated molecules. Lower scores indicate higher estimated synthetic accessibility.}
\label{tab:sa_score}
\centering
\small
\begin{tabular}{@{}lcccc@{}}
\toprule
Method & Mean & Std. & Min & Max \\
\midrule
\textbf{GRPO} & \textbf{3.77} & 0.52 & 3.00 & 5.69 \\
SFT & 4.02 & 0.53 & 3.11 & 5.77 \\
SFT-20x & 3.96 & 0.60 & 2.53 & 5.59 \\
\bottomrule
\end{tabular}
\end{table}

The mean SA score is lowest for GRPO, although this result should be interpreted only as a computational proxy and not as evidence of successful laboratory synthesis.

\begin{table}[h]
\caption{MOSES-style structural-quality metrics evaluated on 50 DFT-validated molecules per method against the 10,000-molecule reference set.}
\label{tab:diversity}
\centering
\small
\begin{tabular}{@{}lccc@{}}
\toprule
Metric & GRPO & SFT & SFT-20x \\
\midrule
Novelty & \textbf{0.980} & 0.940 & 0.920 \\
Scaffold similarity & \textbf{0.154} & 0.295 & 0.284 \\
Fragment similarity & 0.943 & 0.969 & 0.970 \\
SNN & 0.764 & 0.779 & 0.825 \\
Internal diversity & 0.620 & 0.702 & 0.713 \\
Unique scaffolds & \textbf{49/50} & 48/50 & 48/50 \\
\bottomrule
\end{tabular}
\end{table}

GRPO has the lowest scaffold similarity to the reference set and the highest novelty, indicating exploration of new backbone arrangements rather than direct retrieval of known structures. Its fragment similarity remains high, suggesting that the generated molecules reuse chemically plausible building blocks while combining them into novel scaffolds. The lower internal-diversity score than SFT should be interpreted together with the scaffold results: GRPO may reuse successful local motifs while still generating scaffolds that are distant from the reference distribution. Cross-method Tanimoto similarity under matched $(S_1,f)$ targets is approximately 0.31, further indicating that the methods discover structurally distinct solutions.

\section{Limitations}
\label{sec:limitations}

Several limitations remain. First, although the expanded DFT study now contains 20 converged molecules per method, it is still small relative to the size of the OLED chemical space and does not constitute experimental synthesis validation. Second, the BERT predictor is used as a scalable reward proxy during RL, so distribution shift between generated molecules and the labeled dataset remains a potential source of error. The independent DFT evaluation reduces but does not eliminate this concern. Third, the main generation benchmark and the additional GRPO--PPO comparison use different sample sizes (10,000 and 100, respectively); the controlled comparison should therefore be interpreted as an algorithmic ablation rather than a replacement for the larger benchmark. Finally, the current evaluation uses SA scores and scaffold-based statistics as computational proxies for synthesizability and diversity; laboratory feasibility, reaction yield, stability, and device-level OLED performance require future experimental validation.

\section{Conclusion}
\label{sec:conclusion}

We presented a framework for de novo OLED material design that combines a LLaMA-style generative model with Group Relative Policy Optimization (GRPO). By eliminating the need for a value network, the method addresses the critic-estimation challenge in the rugged landscape of chemical properties. The expanded independent DFT evaluation shows that GRPO achieves $S_1$ MAE $=0.260$~eV and $f$ MAE $=0.134$ over 20 converged molecules, while maintaining high generation novelty. In the controlled PPO comparison, GRPO provides better property alignment, and its scaffold-level analysis indicates low similarity to the reference set with plausible fragment usage. At the same time, SFT-20x achieves lower DFT error at the cost of substantial novelty loss, highlighting the central trade-off between property precision and genuinely novel molecular discovery.


\bibliography{references}
\bibliographystyle{unsrt}

\newpage
\appendix

\section{Density Functional Theory (DFT)}
\label{app:dft}

Density Functional Theory (DFT) is a pivotal quantum mechanical computational method for investigating the electronic structure of many-electron systems. Its core tenet is to simplify the $N$-electron wave function $\Psi(\mathbf{r}_1, \mathbf{r}_2, \dots, \mathbf{r}_N)$ to the spatial electron density $\rho(\mathbf{r})$, thereby reducing the dimensionality of the problem from $3N$ to 3. This theory is founded on the Hohenberg-Kohn Theorems: the external potential $V_{\text{ext}}(\mathbf{r})$ is a unique functional of the ground-state density, meaning all ground-state properties of a system are uniquely determined by $\rho(\mathbf{r})$; the theorems also formulate the variational principle for energy, defining the total energy functional $E[\rho]$ of the system, which attains its minimum value at the true ground-state density.

For the practical solution of this functional, Kohn and Sham introduced a non-interacting reference system, decomposing the total energy into the following form:
\begin{equation}
E[\rho] = T_s[\rho] + \int \rho(\mathbf{r})V_{\text{ext}}(\mathbf{r})d\mathbf{r} + \frac{1}{2}\int \frac{\rho(\mathbf{r})\rho(\mathbf{r}')}{|\mathbf{r}-\mathbf{r}'|}d\mathbf{r}d\mathbf{r}' + E_{\text{xc}}[\rho]
\end{equation}
where $T_s[\rho]$ denotes the kinetic energy of non-interacting electrons, the second and third terms represent the nuclear-electron attraction potential energy and the classical electron-electron repulsion energy (Hartree term), respectively. The final term $E_{\text{xc}}[\rho]$ is the exchange-correlation energy, which encapsulates all unknown quantum mechanical many-body effects.

To find the energy minimum, the self-consistent Kohn-Sham equations are derived via the method of variation:
\begin{equation}
\left[ -\frac{1}{2}\nabla^2 + V_{\text{eff}}(\mathbf{r}) \right] \psi_i(\mathbf{r}) = \epsilon_i \psi_i(\mathbf{r})
\end{equation}
in which the effective potential is given by $V_{\text{eff}}(\mathbf{r}) = V_{\text{ext}} + V_H + V_{\text{xc}}$, and the exchange-correlation potential is defined as the functional derivative $V_{\text{xc}}(\mathbf{r}) = \frac{\delta E_{\text{xc}}[\rho]}{\delta \rho(\mathbf{r})}$. Since the exact form of $E_{\text{xc}}$ remains unknown, approximate schemes such as the Local Density Approximation (LDA) and Generalized Gradient Approximation (GGA) are commonly adopted in practical applications.

The entire calculation process is implemented iteratively via the Self-Consistent Field (SCF) method until the electron density or total energy meets the predefined convergence criterion. Endowed with excellent computational efficiency and reliable accuracy for handling large atomic systems, DFT has become an indispensable tool in computational chemistry and computational materials science.

\section{DFT Validation Sample Selection}
\label{app:dft_samples}

\begin{figure}[h]
\centering
\includegraphics[width=\textwidth]{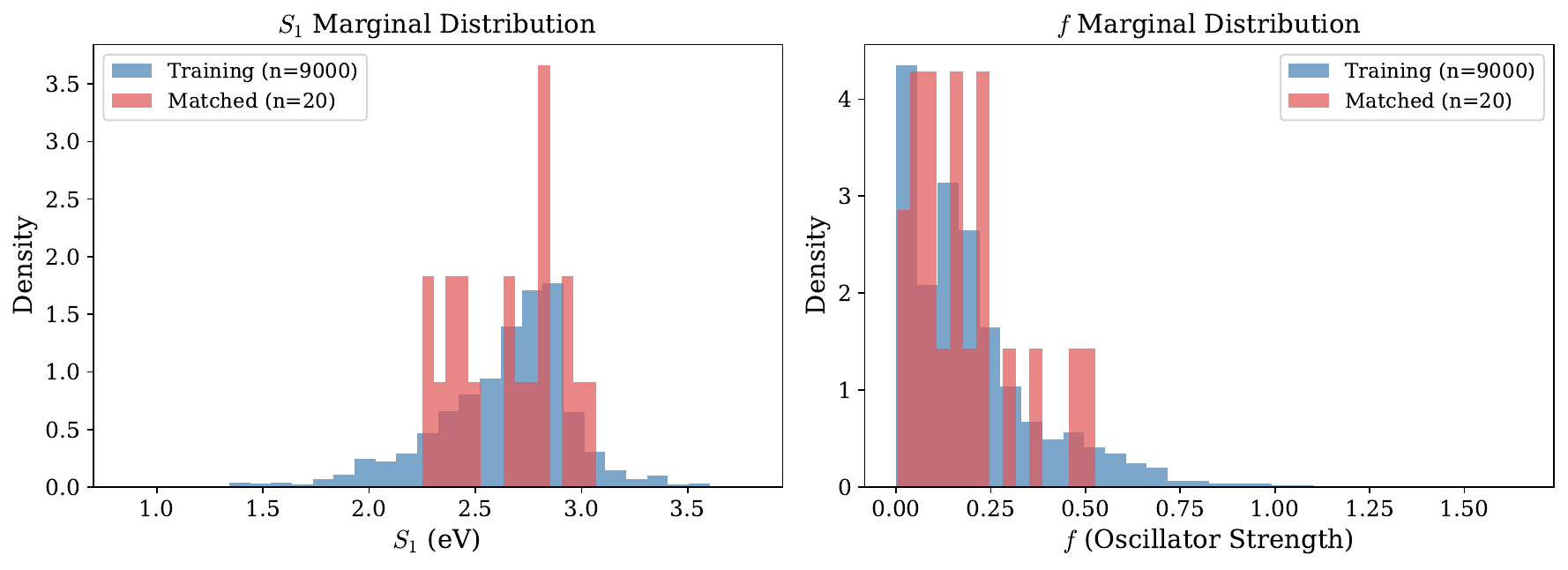}
\caption{Marginal distributions of target properties. The training set distribution (blue, n=9000) is compared with the matched DFT validation samples (red, n=20). The selected samples cover the high-density regions of the training distribution while ensuring diversity in both $S_1$ and oscillator strength $f$.}
\label{fig:marginal_dist}
\end{figure}

The 20 target property pairs for DFT validation were selected to cover diverse regions of the training distribution. Figure~\ref{fig:marginal_dist} shows the marginal distributions of $S_1$ and $f$ for both the training set and the matched samples, demonstrating comprehensive coverage of the property space.

\section{RDKit Cheminformatics Platform}
\label{app:rdkit}

RDKit is an open-source computational platform dedicated to molecular modeling and cheminformatics processing. It abstracts chemical molecules as attributed graphs $G = (V, E)$ consisting of vertices (atoms) and edges (chemical bonds). In processing molecular structures, RDKit parses standardized representations such as SMILES and SDF, converting chemical topological structures into machine-processable adjacency matrices and atomic attribute vectors. RDKit features three core functional modules as follows:

\paragraph{Molecular Fingerprinting.} To enable the integration of chemical structures with machine learning algorithms, RDKit provides a variety of molecular fingerprinting algorithms, with the hash-based Morgan fingerprint being the most representative. This algorithm iteratively collects information on atoms and their local neighborhood environments, mapping the local chemical environment to a fixed-length binary vector $\mathbf{V}$. This representation can capture the substructural features of molecules, and the similarity between molecules is quantified by the Tanimoto coefficient $T_c$, thus facilitating chemical space clustering and similarity screening.

\paragraph{Conformational Space Search and Force Field Optimization.} For studies involving three-dimensional molecular properties, RDKit uses the Distance Geometry method to randomly generate initial 3D coordinates of molecules within the bounds of constraints such as bond lengths and bond angles. Subsequently, RDKit invokes built-in force fields (e.g., the Universal Force Field (UFF) or Merck Molecular Force Field 94 (MMFF94)) to perform energy minimization on the potential energy surface, providing high-quality initial structures for high-level quantum chemical calculations.

\paragraph{Chemical Descriptor Calculation.} In addition to topological features, RDKit can calculate more than 200 physicochemical descriptors, such as the octanol-water partition coefficient (logP), molar refractivity, Topological Polar Surface Area (TPSA), and molecular weight. These parameters are obtained via weighted summation using empirical formulas and Group Contribution Methods, serving as key indicators for evaluating the pharmacokinetic properties of molecules.

\section{Scientific Analysis of Generated Molecules}
\label{app:mol_viewer}

\paragraph{This section presents representative OLED molecules generated by our GRPO-aligned model across different property buckets. Each figure shows the 2D structure (top), 3D conformation (bottom), target properties, and SMILES representation.}

From a chemical perspective, the structures generated exhibit the following characteristics:
\begin{enumerate}
    \item Among the 50 OLED molecules generated within the $S_1 \in [2.5, 3]$ eV interval, there are 10 with B/N/O backbones, 1 with a B/O/O backbone, and the remaining 39 are B/N/N backbones, accounting for approximately 80\% of the total.
    \item The generated structures include both classical mono-MR frameworks with symmetrically substituted groups (e.g., \texttt{rl\_bucket1\_idx2}, \texttt{rl\_bucket2\_idx0}, and \texttt{rl\_bucket3\_idx2}), as well as large conjugated systems extended through fusion with heteroatoms such as N or S based on mono-MR frameworks (e.g., \texttt{rl\_bucket5\_idx4}, \texttt{rl\_bucket2\_idx6}, and \texttt{rl\_bucket3\_idx0}).
\end{enumerate}

The corresponding molecular SMILES, 2D structures, and 3D conformations are presented as follows.

\begin{figure}[htbp]
\centering
\begin{minipage}{0.32\textwidth}
    \centering
    \includegraphics[width=\textwidth]{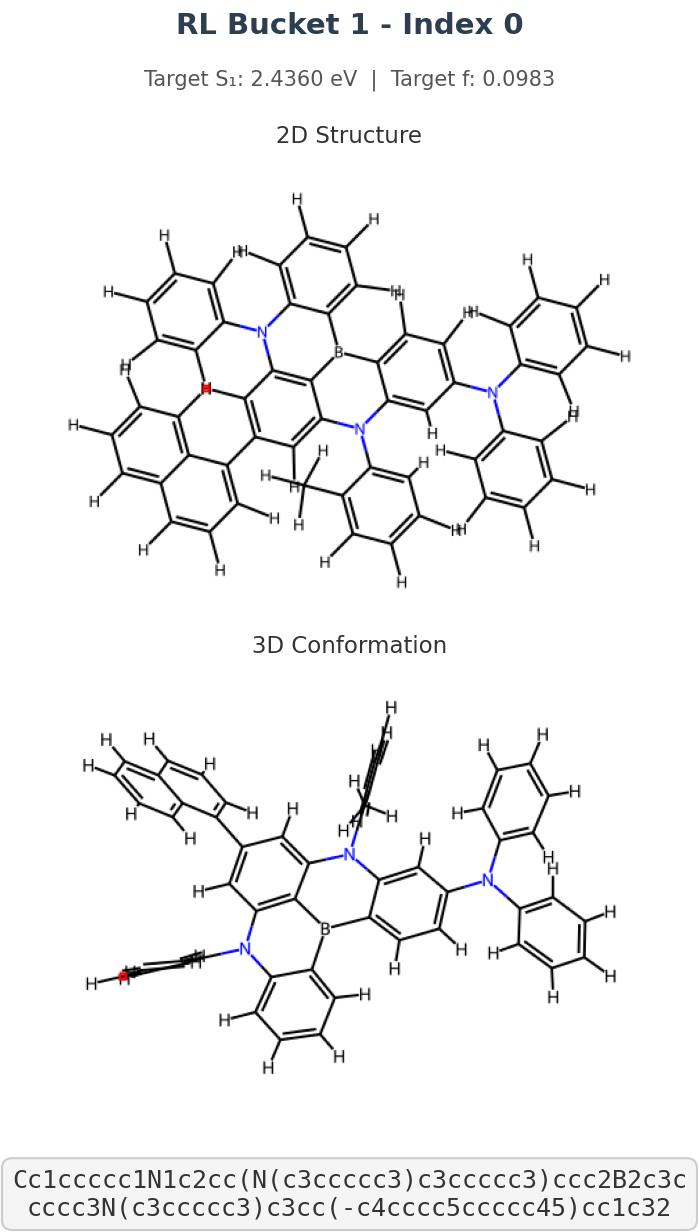}
\end{minipage}
\begin{minipage}{0.32\textwidth}
    \centering
    \includegraphics[width=\textwidth]{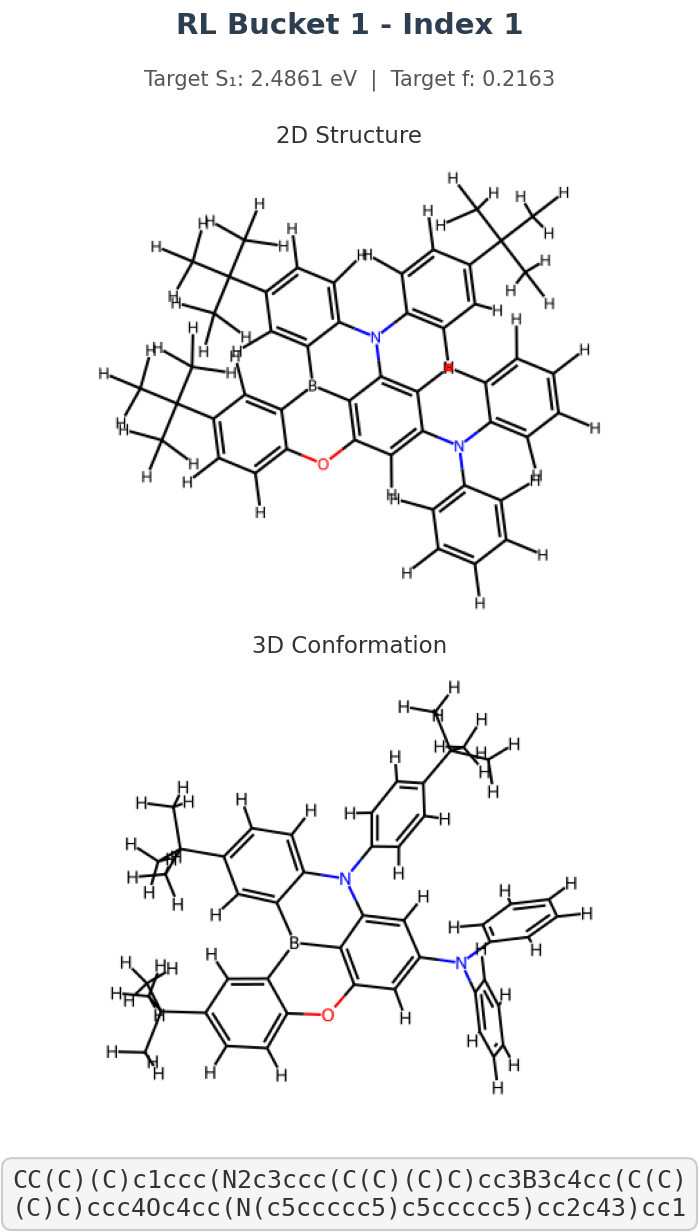}
\end{minipage}
\begin{minipage}{0.32\textwidth}
    \centering
    \includegraphics[width=\textwidth]{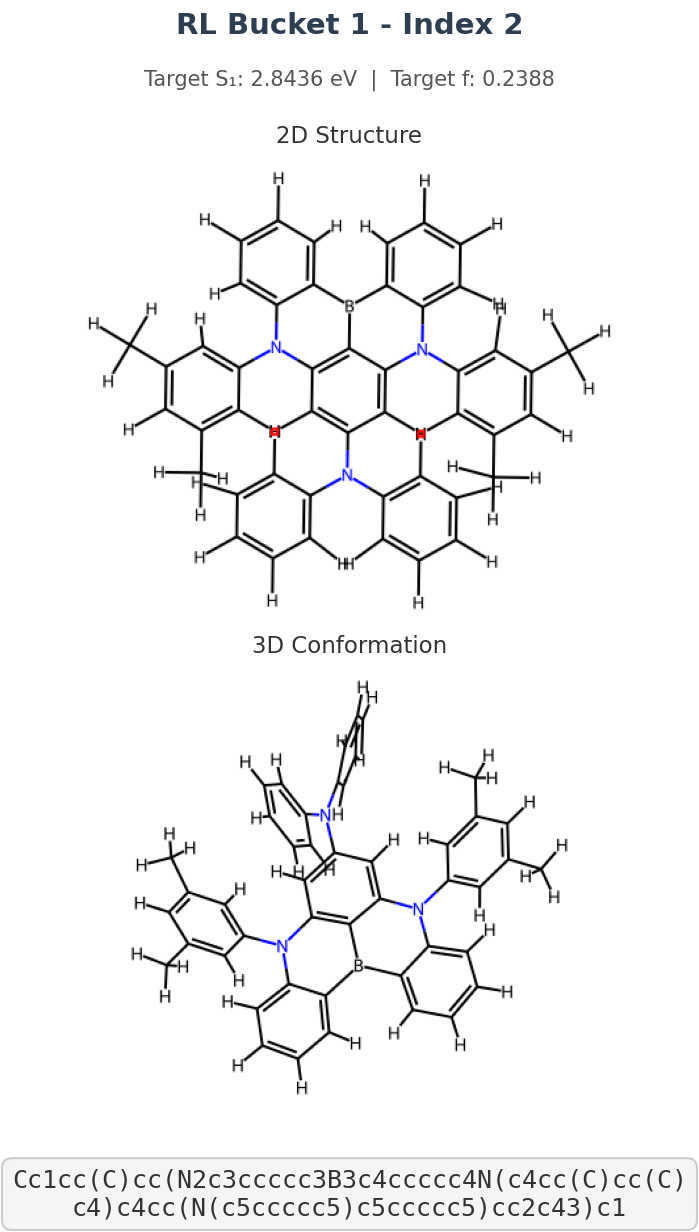}
\end{minipage}
\caption{Generated OLED molecules from Bucket 1 (1-3).}
\label{fig:bucket1a}
\end{figure}

\begin{figure}[htbp]
\centering
\begin{minipage}{0.32\textwidth}
    \centering
    \includegraphics[width=\textwidth]{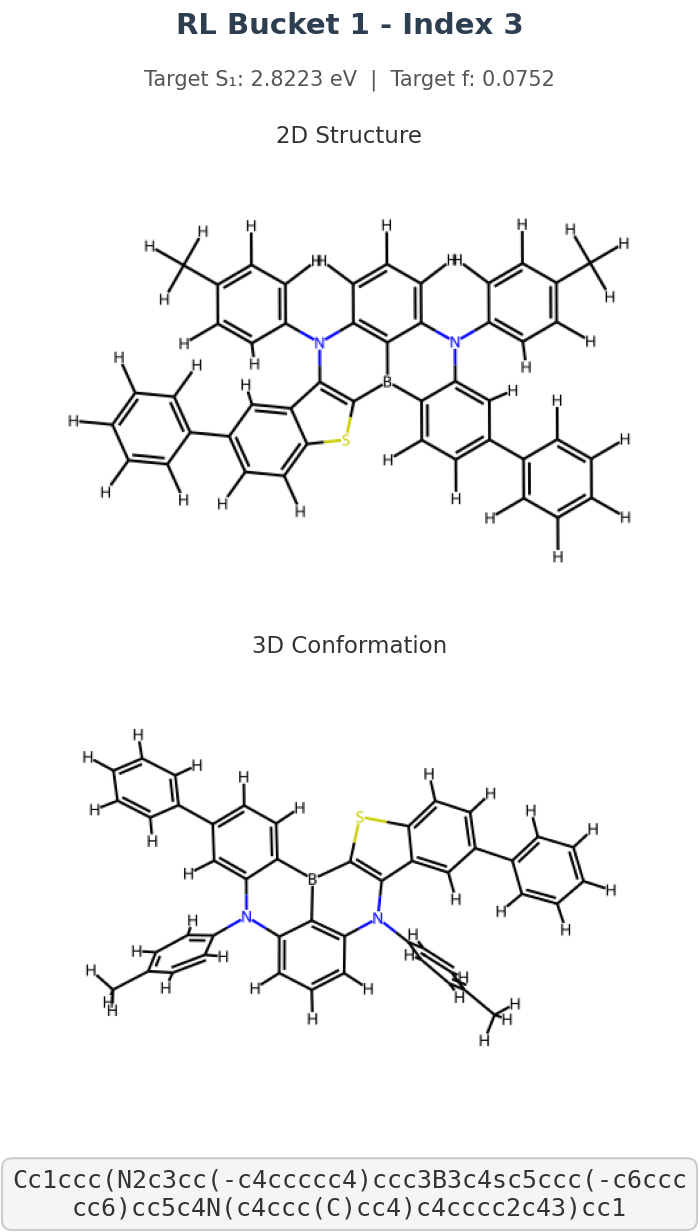}
\end{minipage}
\begin{minipage}{0.32\textwidth}
    \centering
    \includegraphics[width=\textwidth]{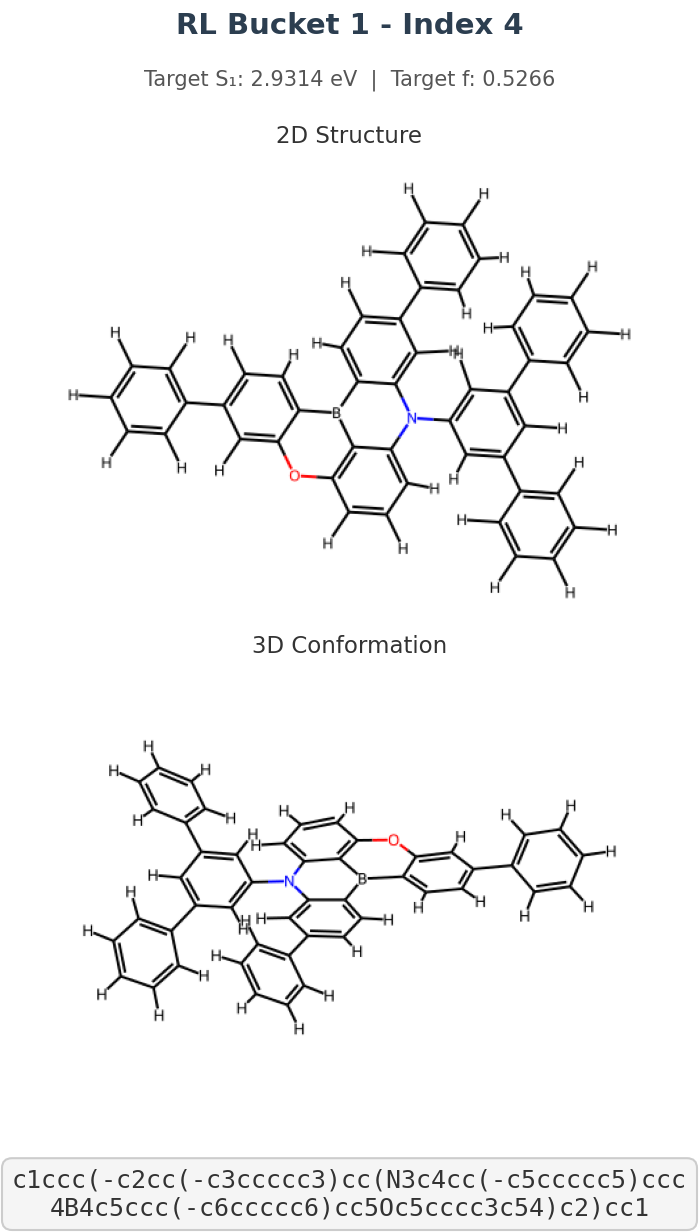}
\end{minipage}
\begin{minipage}{0.32\textwidth}
    \centering
    \includegraphics[width=\textwidth]{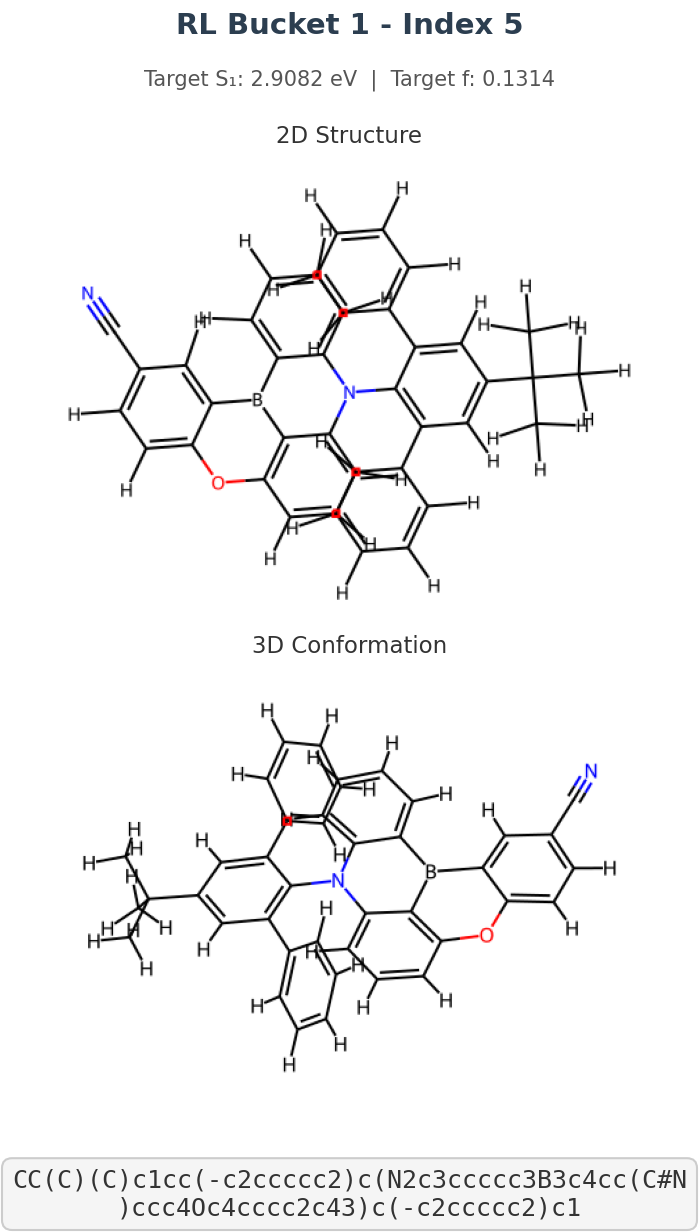}
\end{minipage}
\caption{Generated OLED molecules from Bucket 1 (4-6).}
\label{fig:bucket1b}
\end{figure}

\begin{figure}[htbp]
\centering
\begin{minipage}{0.32\textwidth}
    \centering
    \includegraphics[width=\textwidth]{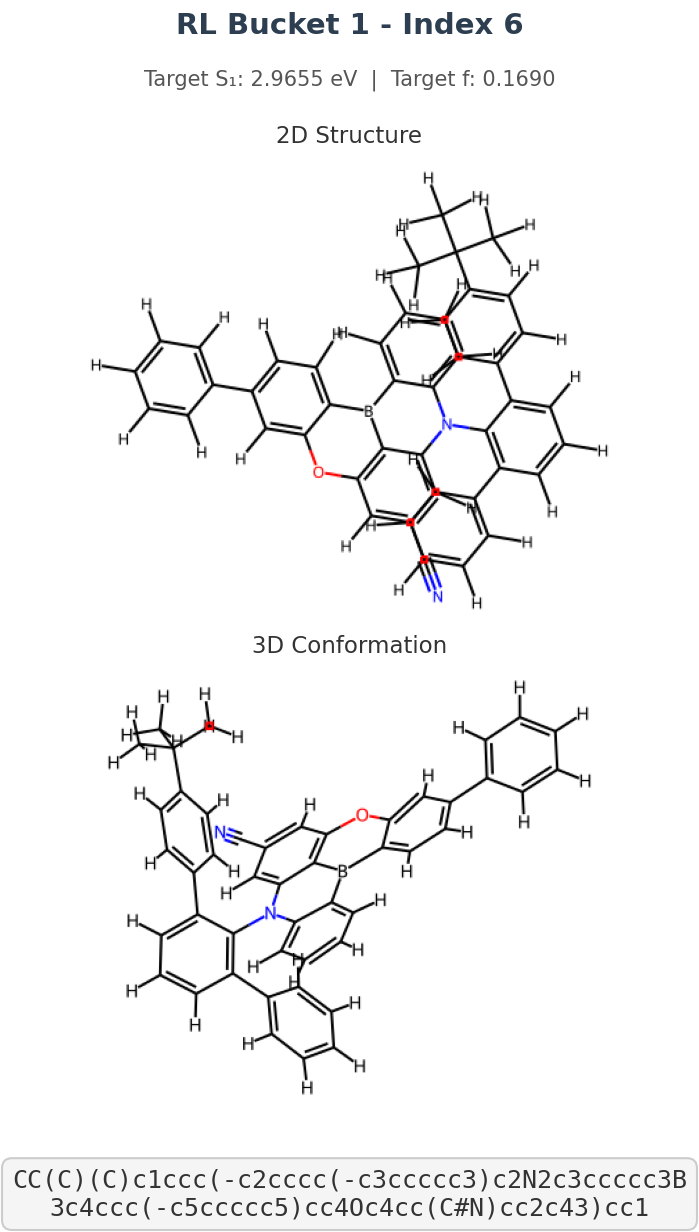}
\end{minipage}
\begin{minipage}{0.32\textwidth}
    \centering
    \includegraphics[width=\textwidth]{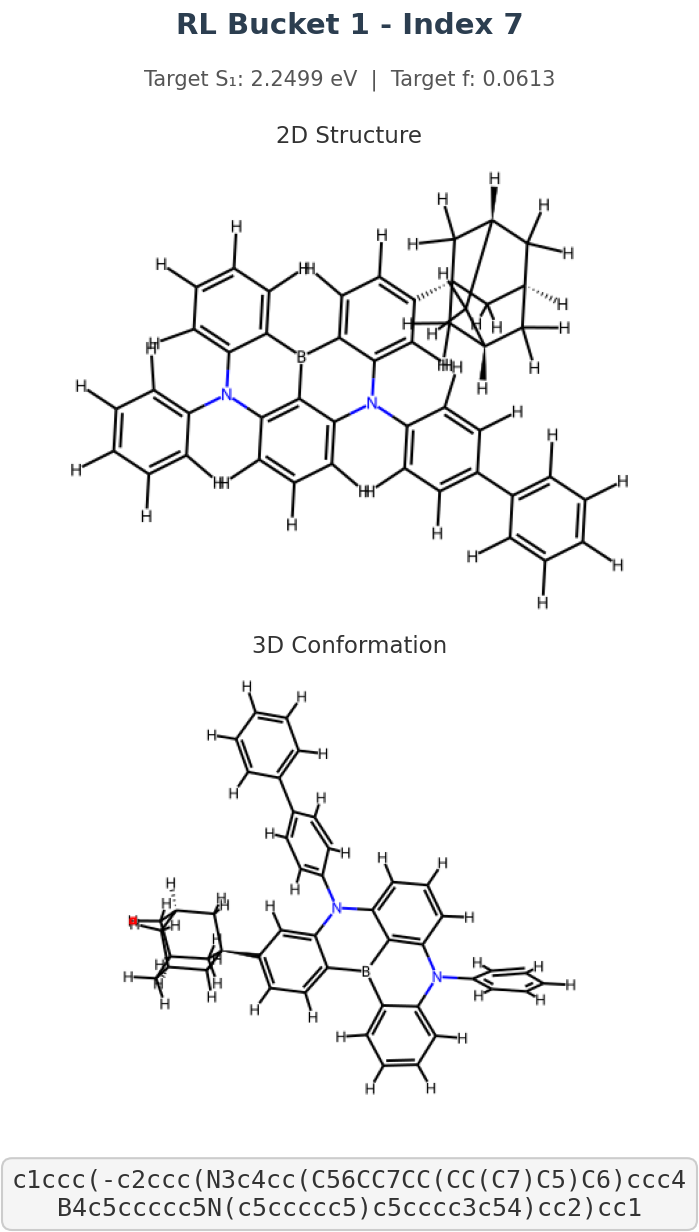}
\end{minipage}
\begin{minipage}{0.32\textwidth}
    \centering
    \includegraphics[width=\textwidth]{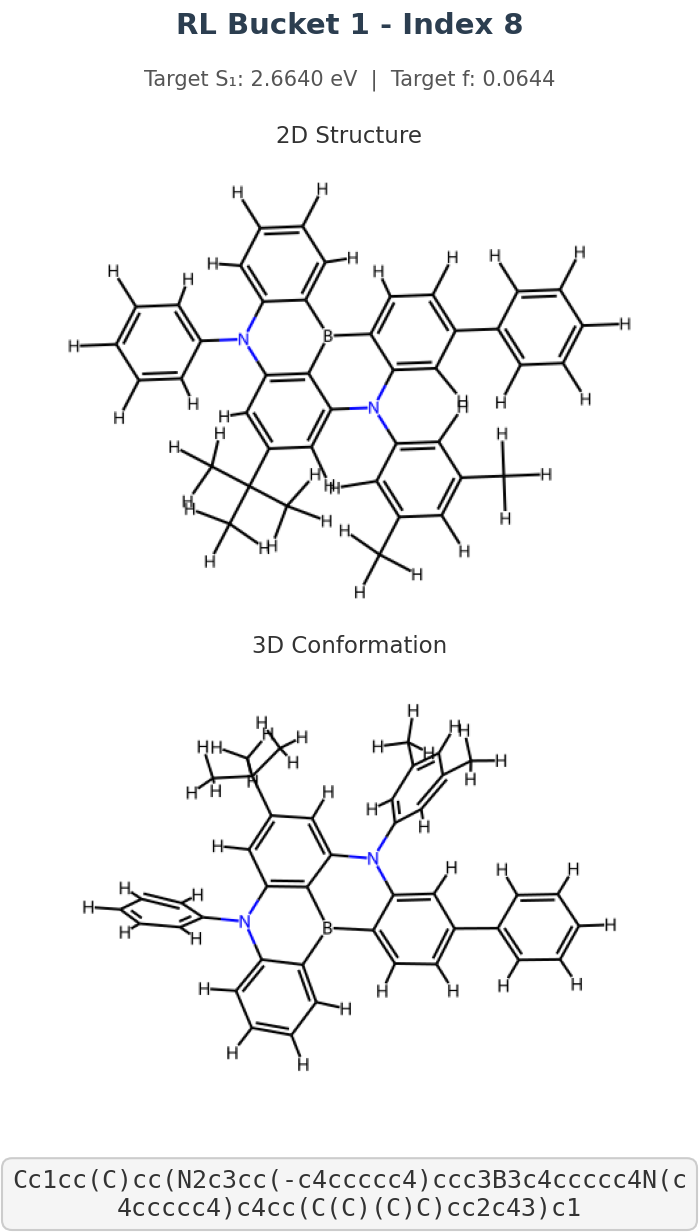}
\end{minipage}
\caption{Generated OLED molecules from Bucket 1 (7-9).}
\label{fig:bucket1c}
\end{figure}

\begin{figure}[htbp]
\centering
\begin{minipage}{0.32\textwidth}
    \centering
    \includegraphics[width=\textwidth]{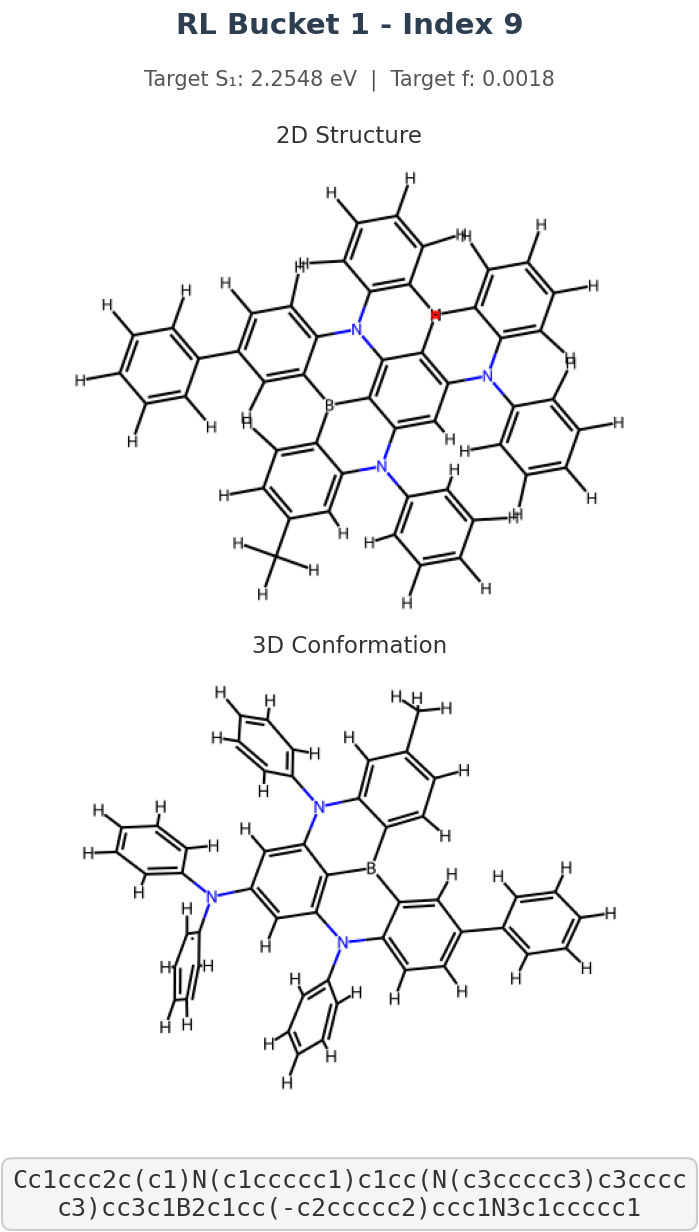}
\end{minipage}
\caption{Generated OLED molecules from Bucket 1 (10).}
\label{fig:bucket1d}
\end{figure}

\begin{figure}[htbp]
\centering
\begin{minipage}{0.32\textwidth}
    \centering
    \includegraphics[width=\textwidth]{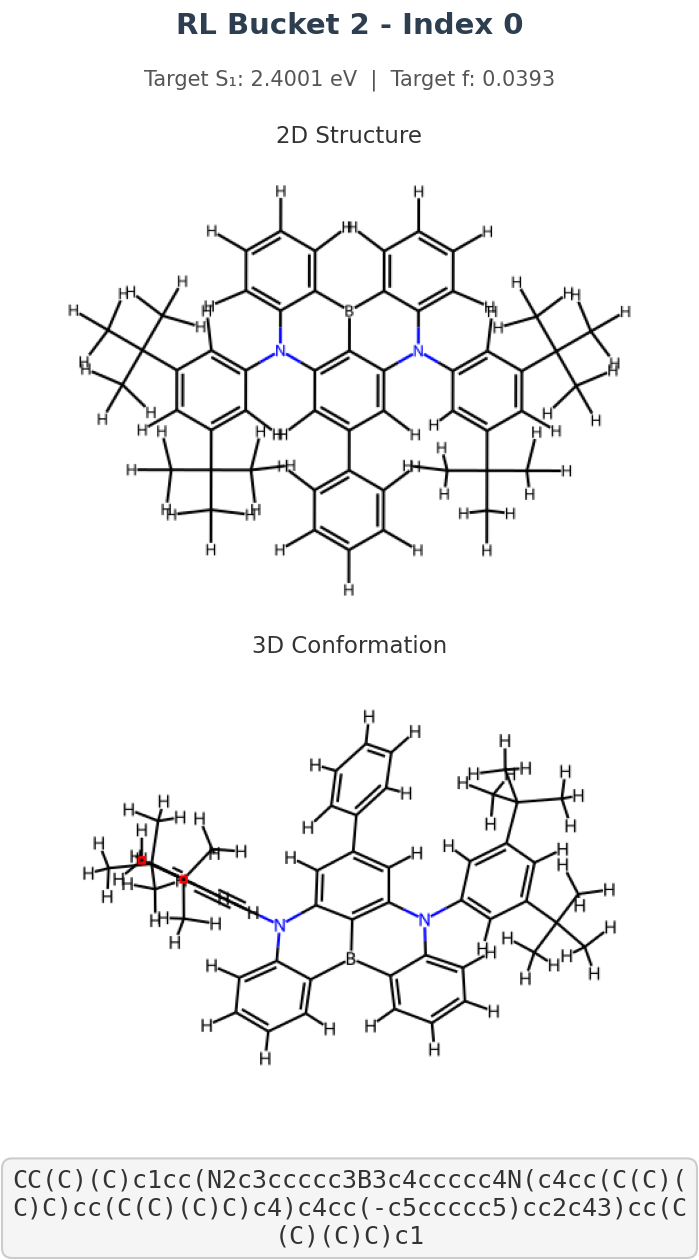}
\end{minipage}
\begin{minipage}{0.32\textwidth}
    \centering
    \includegraphics[width=\textwidth]{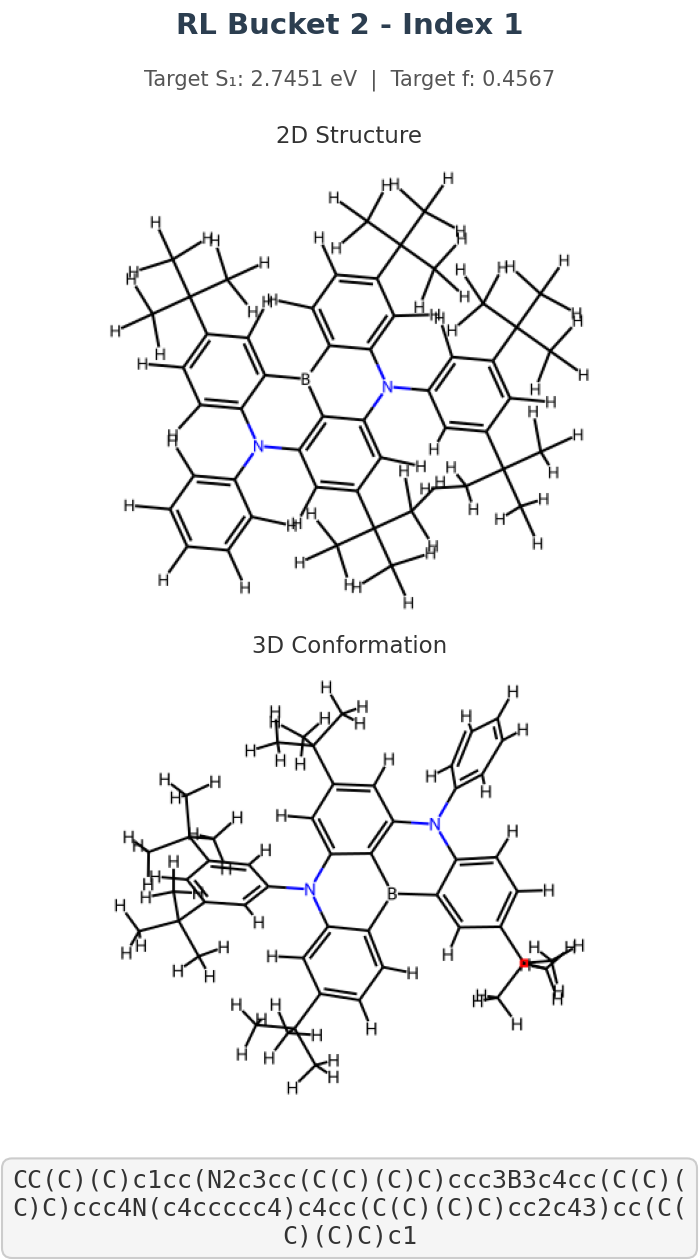}
\end{minipage}
\begin{minipage}{0.32\textwidth}
    \centering
    \includegraphics[width=\textwidth]{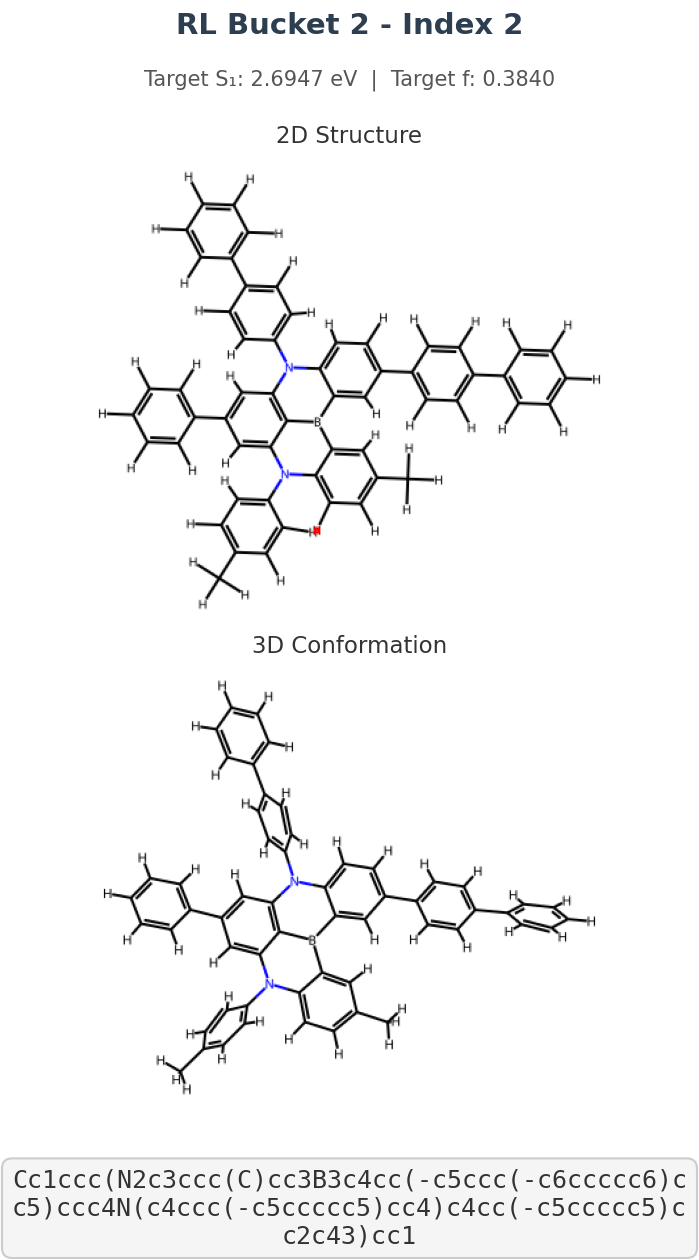}
\end{minipage}
\caption{Generated OLED molecules from Bucket 2 (1-3).}
\label{fig:bucket2a}
\end{figure}

\begin{figure}[htbp]
\centering
\begin{minipage}{0.32\textwidth}
    \centering
    \includegraphics[width=\textwidth]{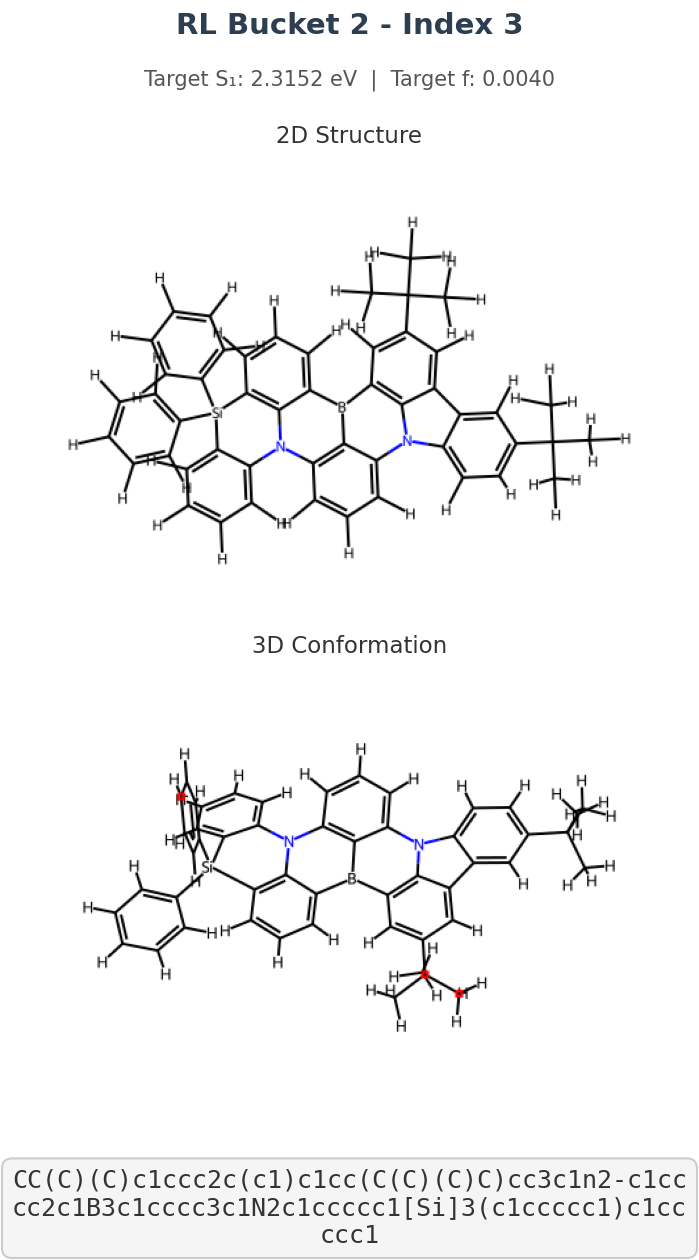}
\end{minipage}
\begin{minipage}{0.32\textwidth}
    \centering
    \includegraphics[width=\textwidth]{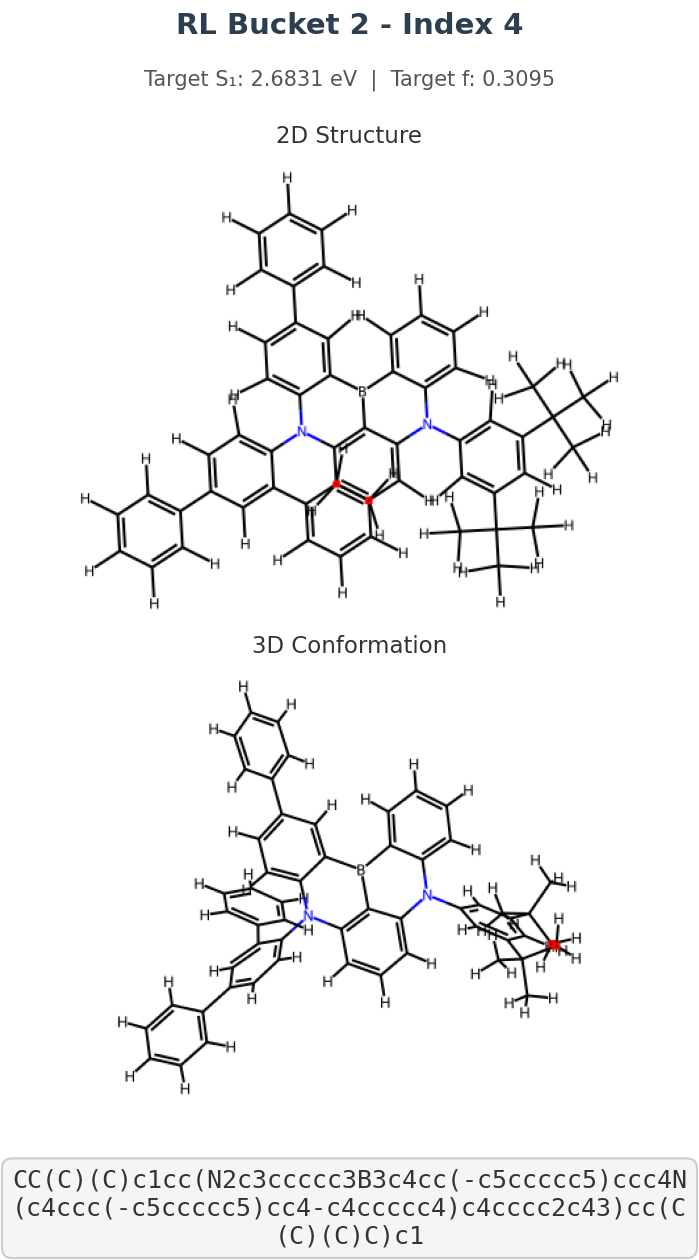}
\end{minipage}
\begin{minipage}{0.32\textwidth}
    \centering
    \includegraphics[width=\textwidth]{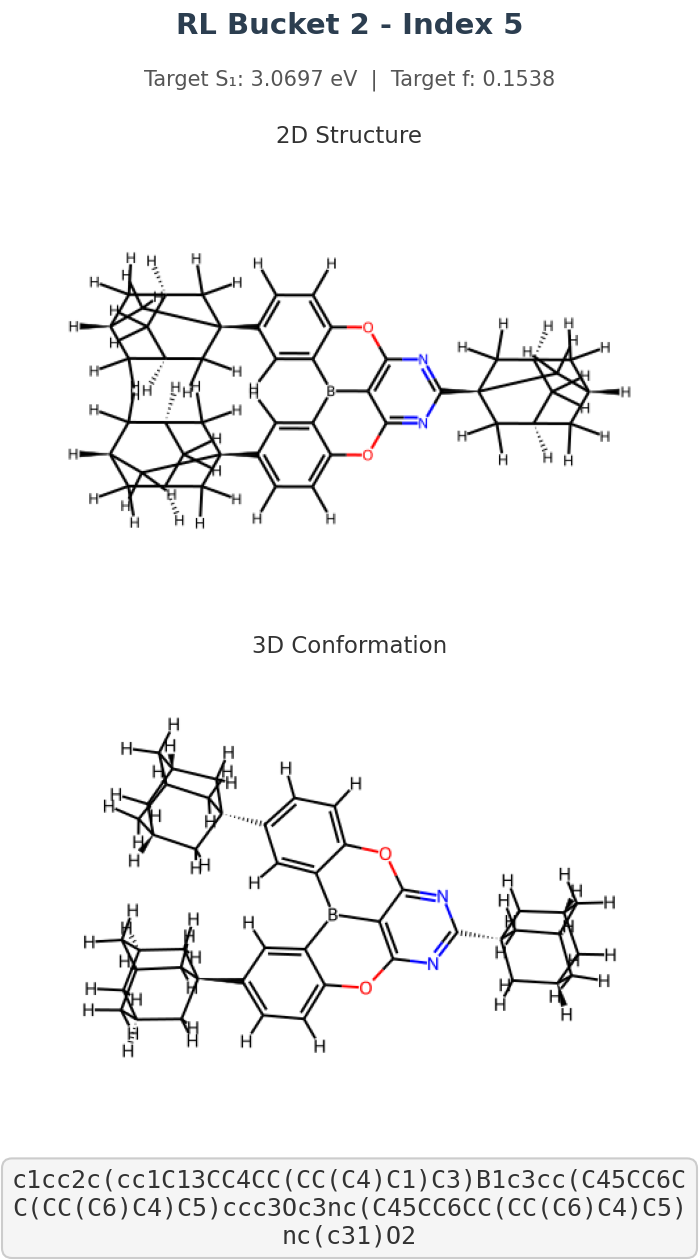}
\end{minipage}
\caption{Generated OLED molecules from Bucket 2 (4-6).}
\label{fig:bucket2b}
\end{figure}

\begin{figure}[htbp]
\centering
\begin{minipage}{0.32\textwidth}
    \centering
    \includegraphics[width=\textwidth]{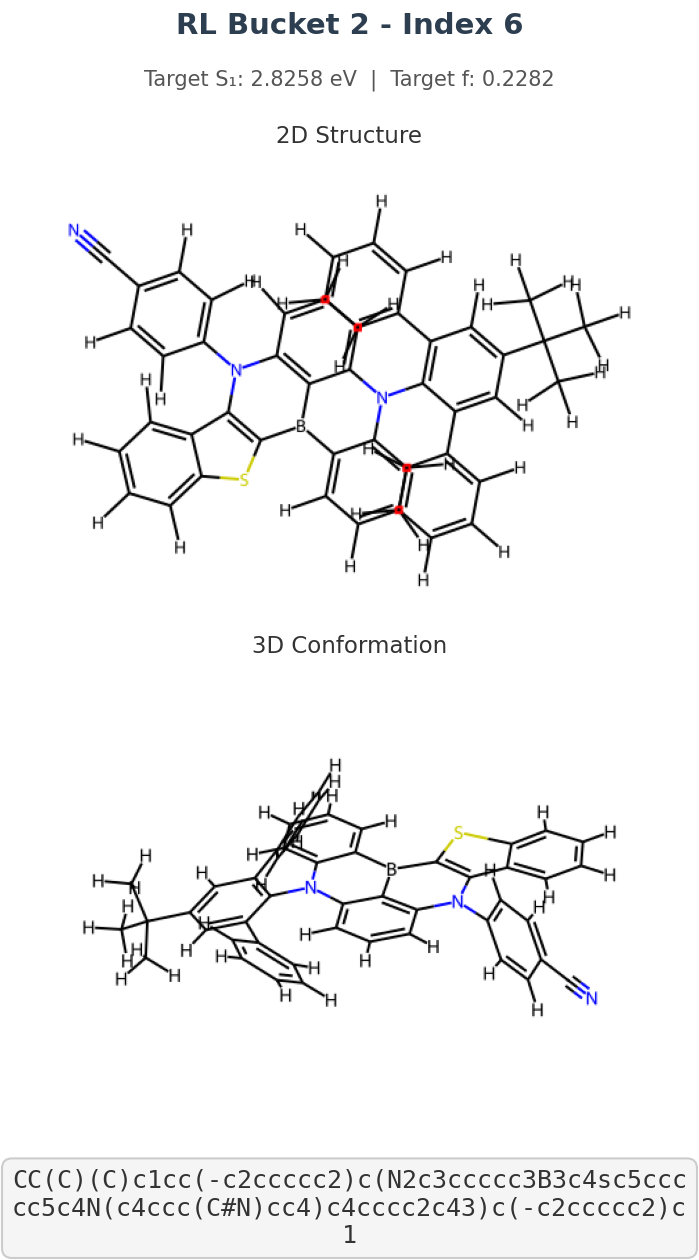}
\end{minipage}
\begin{minipage}{0.32\textwidth}
    \centering
    \includegraphics[width=\textwidth]{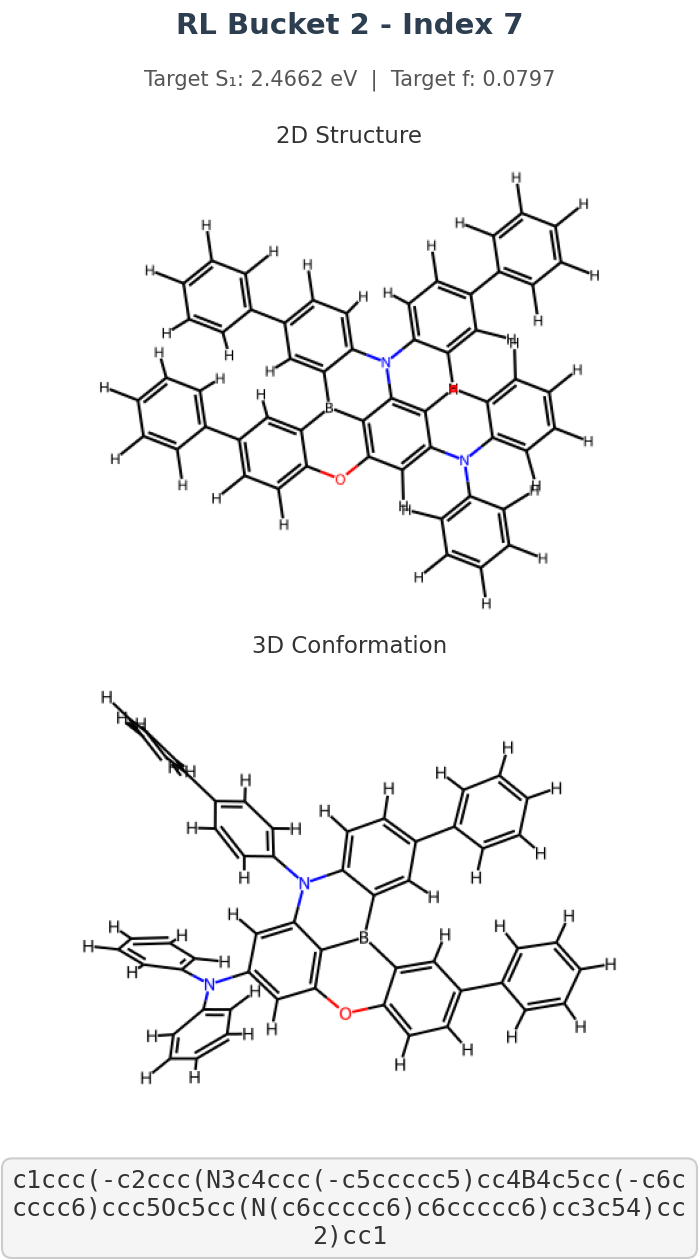}
\end{minipage}
\begin{minipage}{0.32\textwidth}
    \centering
    \includegraphics[width=\textwidth]{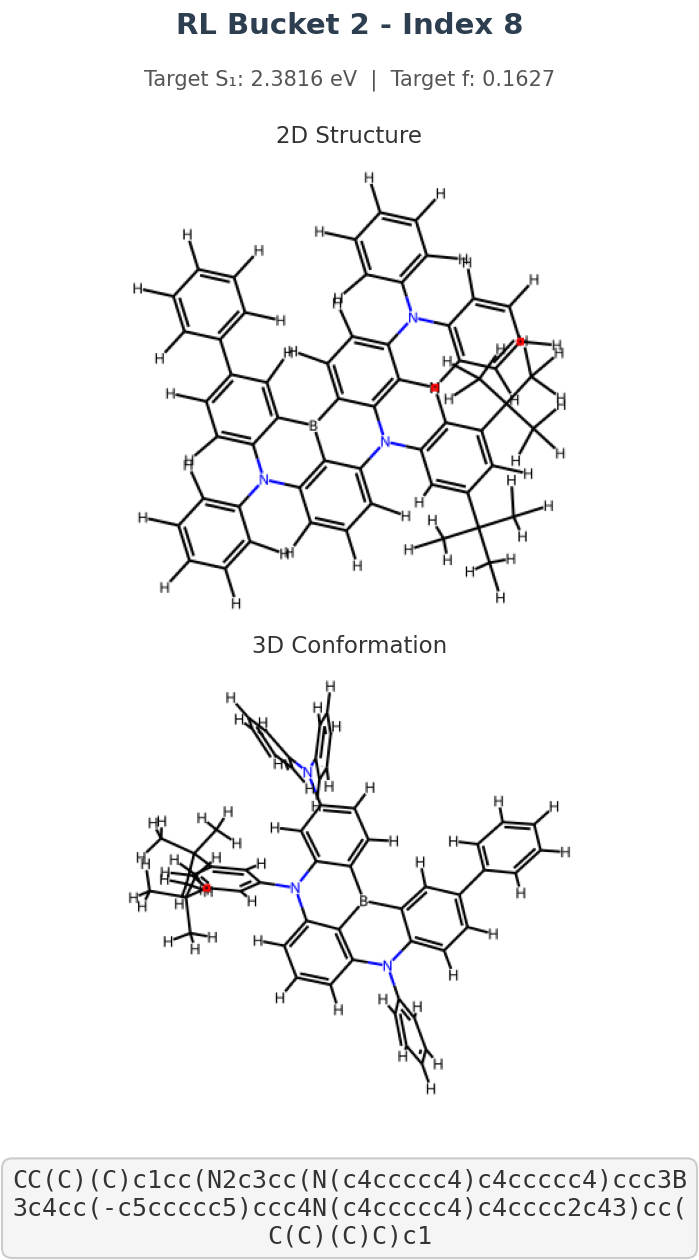}
\end{minipage}
\caption{Generated OLED molecules from Bucket 2 (7-9).}
\label{fig:bucket2c}
\end{figure}

\begin{figure}[htbp]
\centering
\begin{minipage}{0.32\textwidth}
    \centering
    \includegraphics[width=\textwidth]{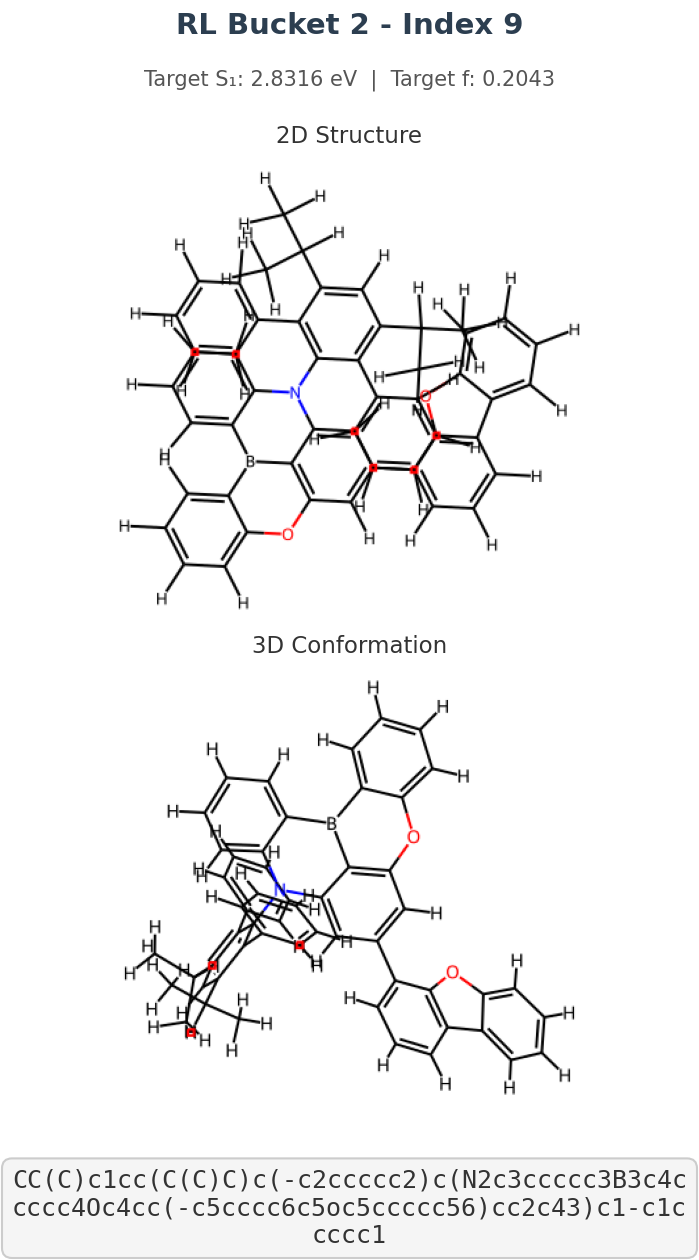}
\end{minipage}
\caption{Generated OLED molecules from Bucket 2 (10).}
\label{fig:bucket2d}
\end{figure}

\begin{figure}[htbp]
\centering
\begin{minipage}{0.32\textwidth}
    \centering
    \includegraphics[width=\textwidth]{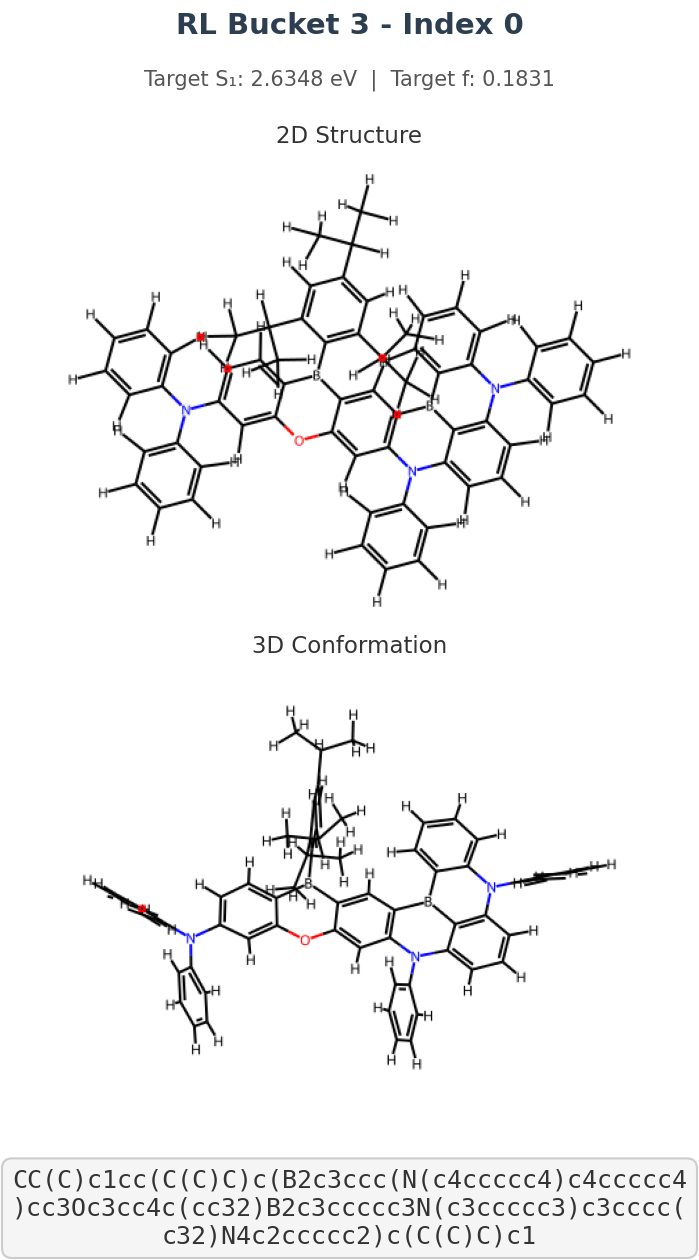}
\end{minipage}
\begin{minipage}{0.32\textwidth}
    \centering
    \includegraphics[width=\textwidth]{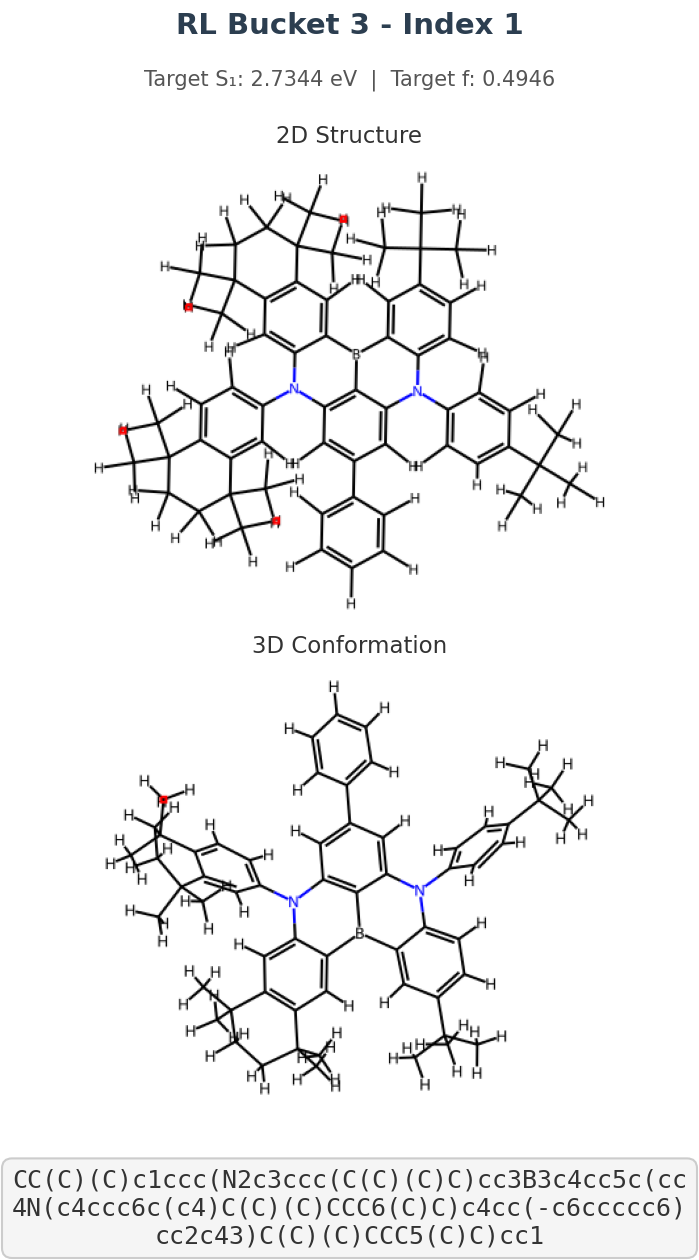}
\end{minipage}
\begin{minipage}{0.32\textwidth}
    \centering
    \includegraphics[width=\textwidth]{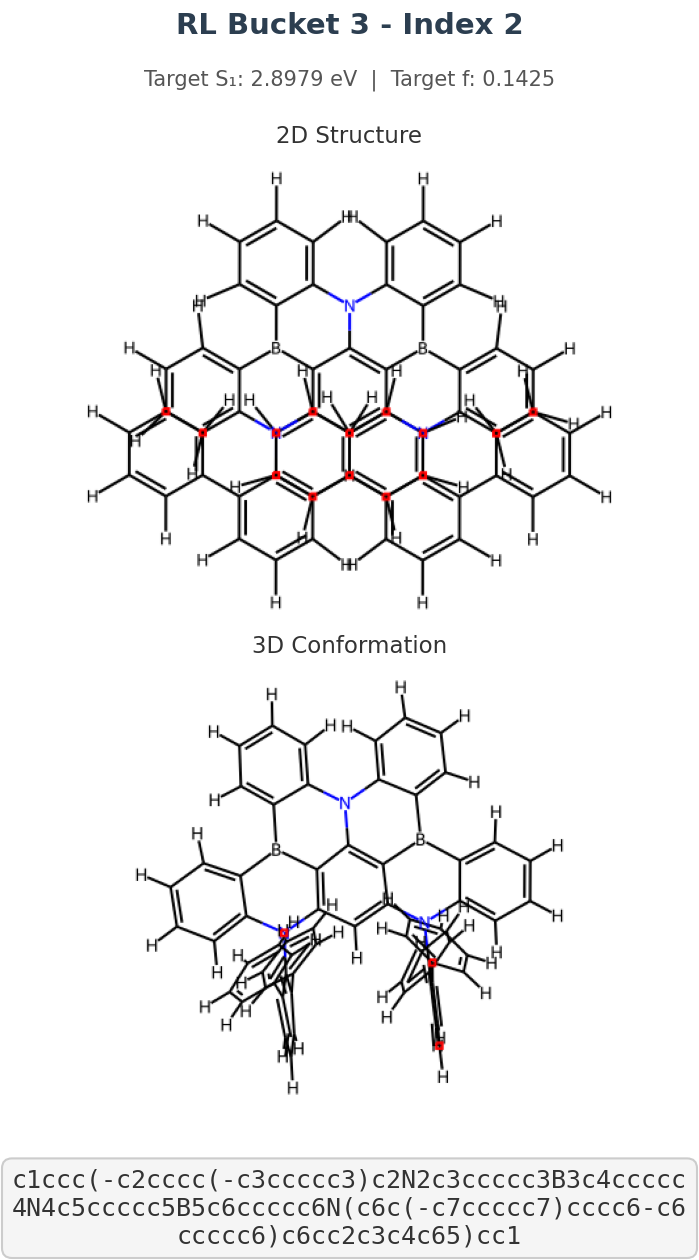}
\end{minipage}
\caption{Generated OLED molecules from Bucket 3 (1-3).}
\label{fig:bucket3a}
\end{figure}

\begin{figure}[htbp]
\centering
\begin{minipage}{0.32\textwidth}
    \centering
    \includegraphics[width=\textwidth]{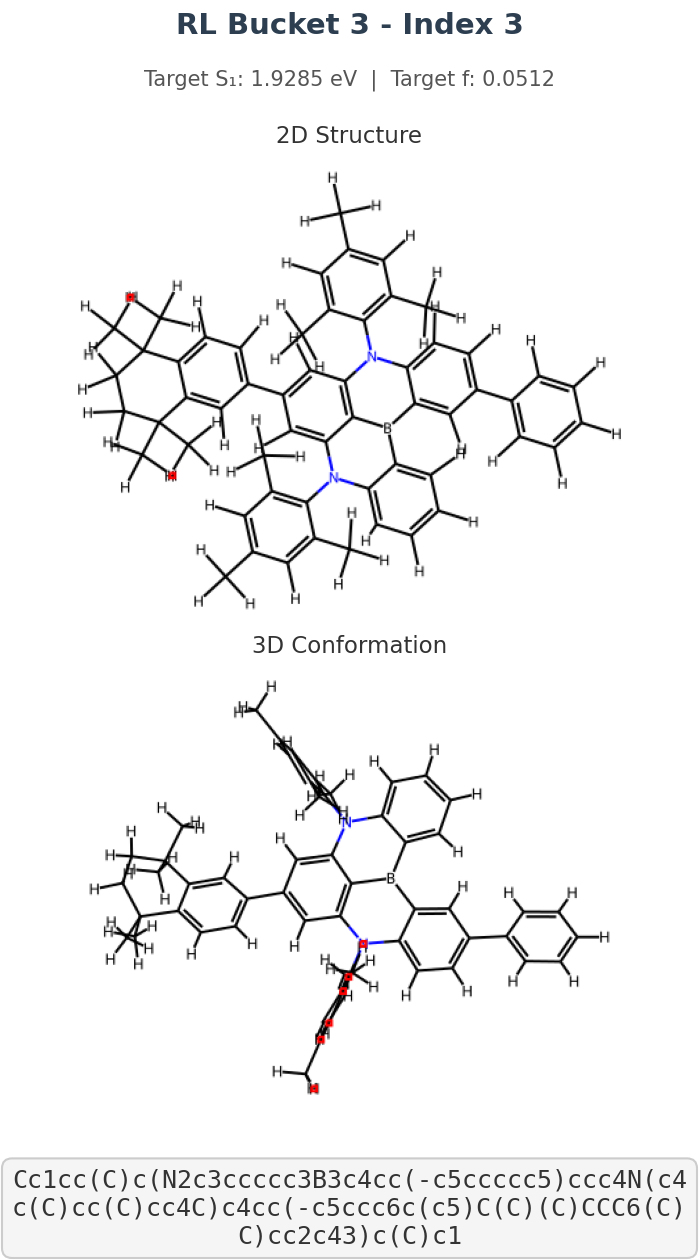}
\end{minipage}
\begin{minipage}{0.32\textwidth}
    \centering
    \includegraphics[width=\textwidth]{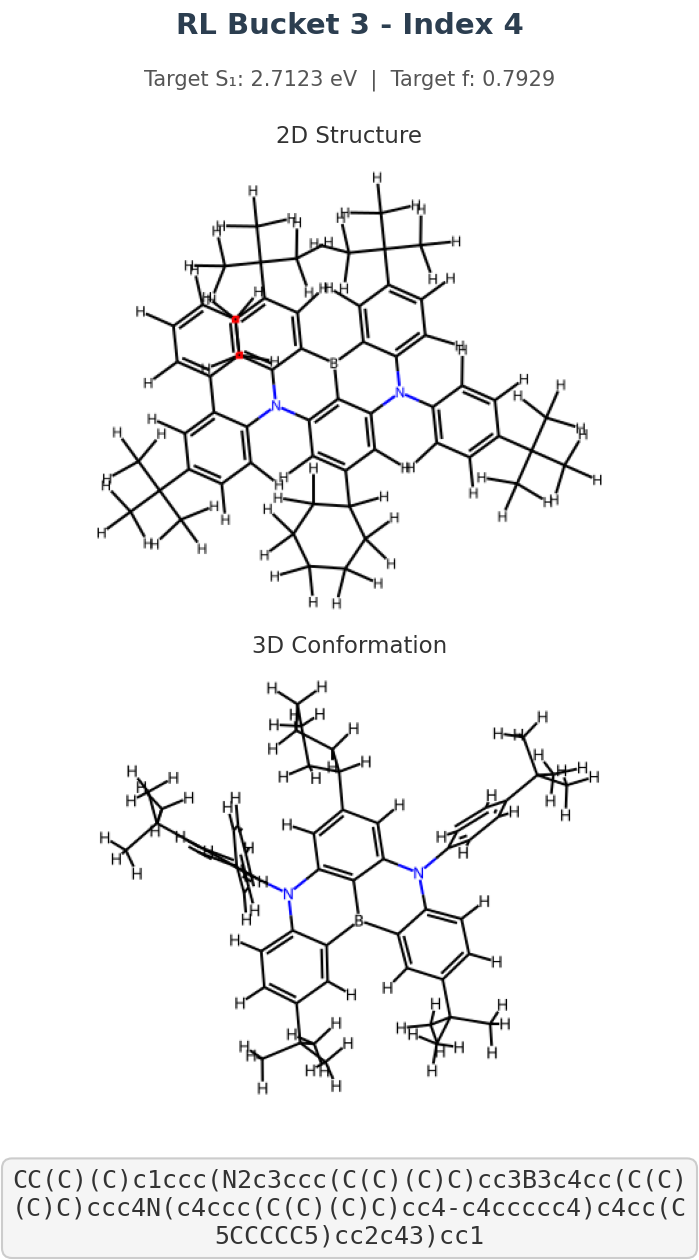}
\end{minipage}
\begin{minipage}{0.32\textwidth}
    \centering
    \includegraphics[width=\textwidth]{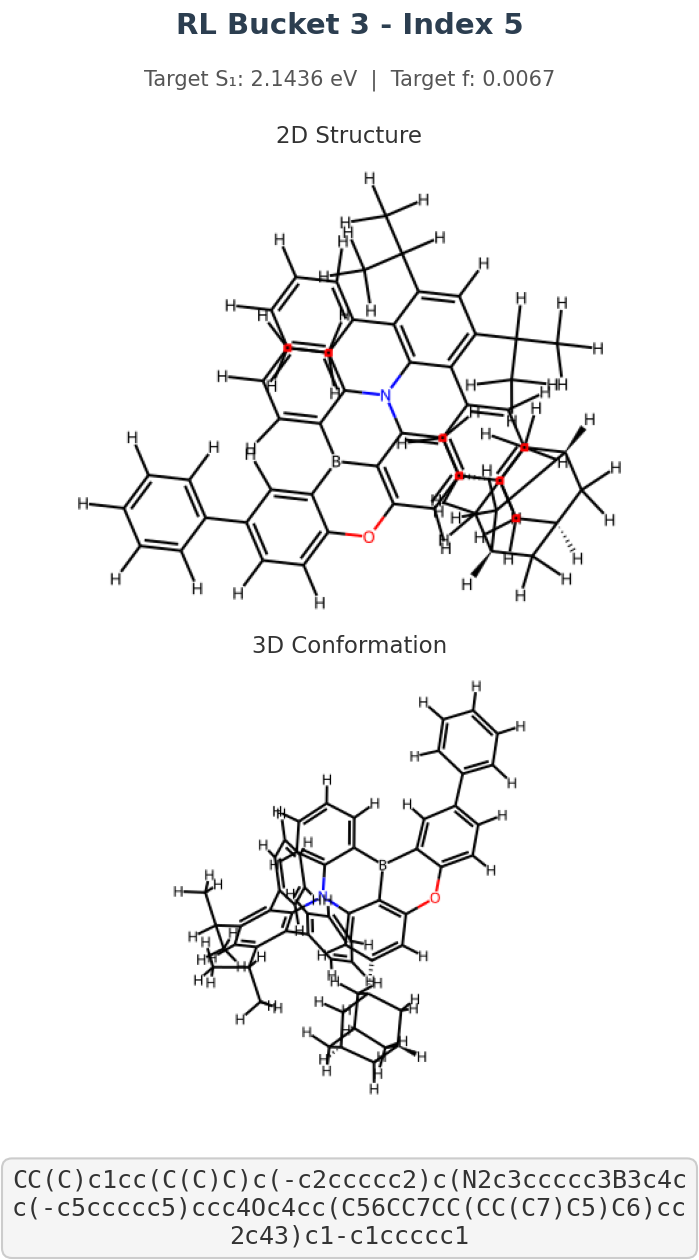}
\end{minipage}
\caption{Generated OLED molecules from Bucket 3 (4-6).}
\label{fig:bucket3b}
\end{figure}

\begin{figure}[htbp]
\centering
\begin{minipage}{0.32\textwidth}
    \centering
    \includegraphics[width=\textwidth]{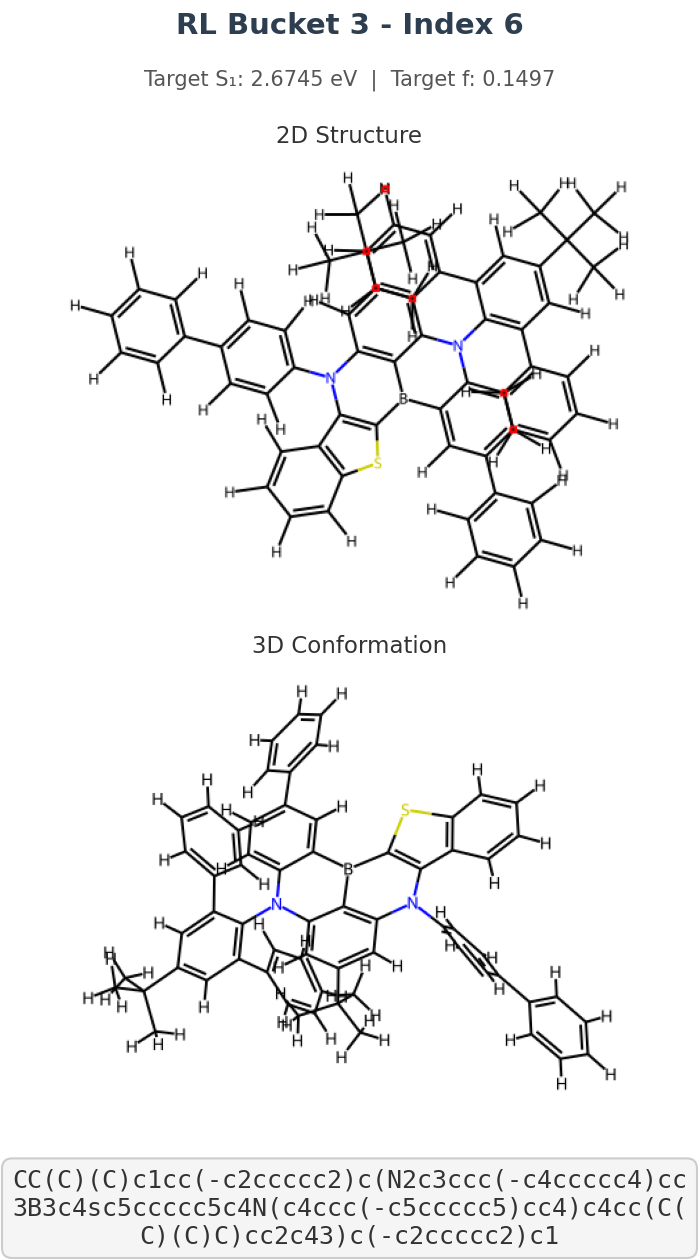}
\end{minipage}
\begin{minipage}{0.32\textwidth}
    \centering
    \includegraphics[width=\textwidth]{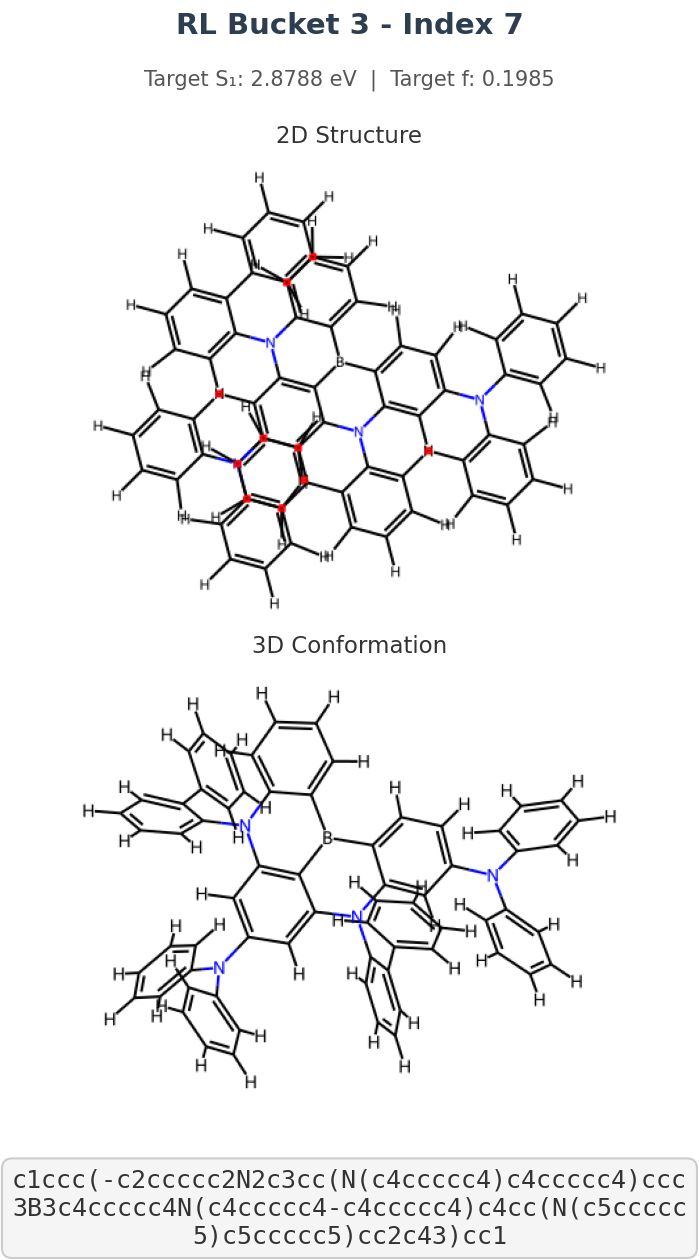}
\end{minipage}
\begin{minipage}{0.32\textwidth}
    \centering
    \includegraphics[width=\textwidth]{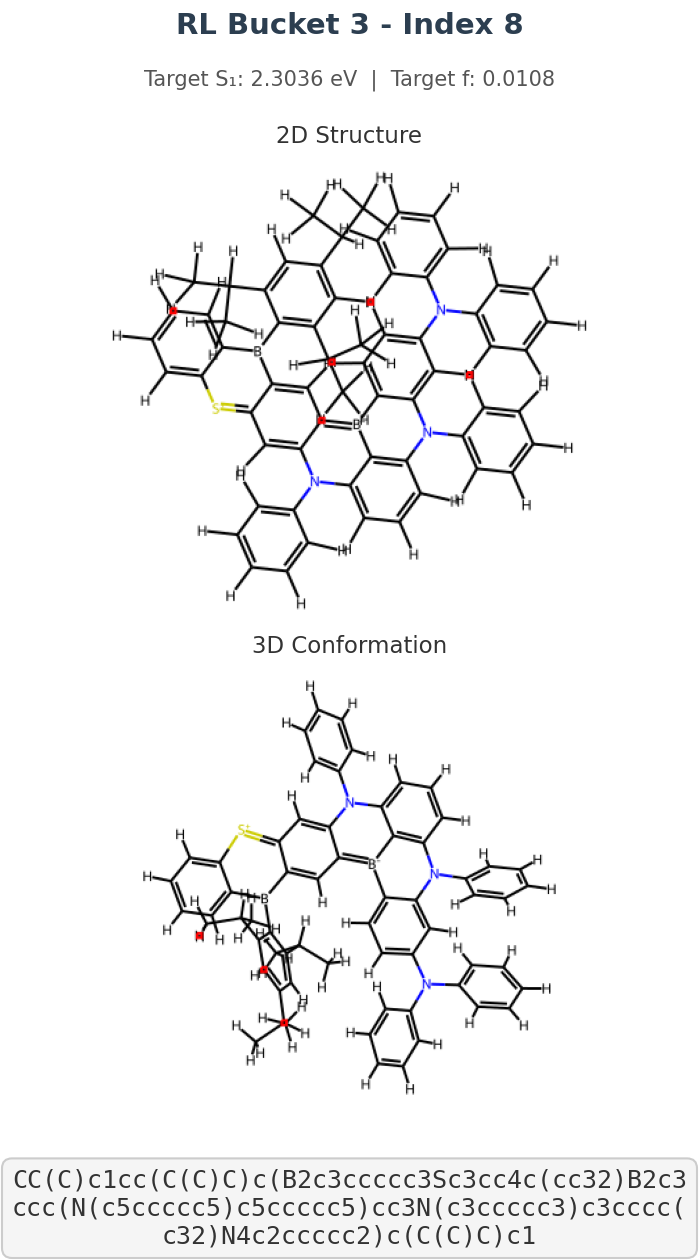}
\end{minipage}
\caption{Generated OLED molecules from Bucket 3 (7-9).}
\label{fig:bucket3c}
\end{figure}

\begin{figure}[htbp]
\centering
\begin{minipage}{0.32\textwidth}
    \centering
    \includegraphics[width=\textwidth]{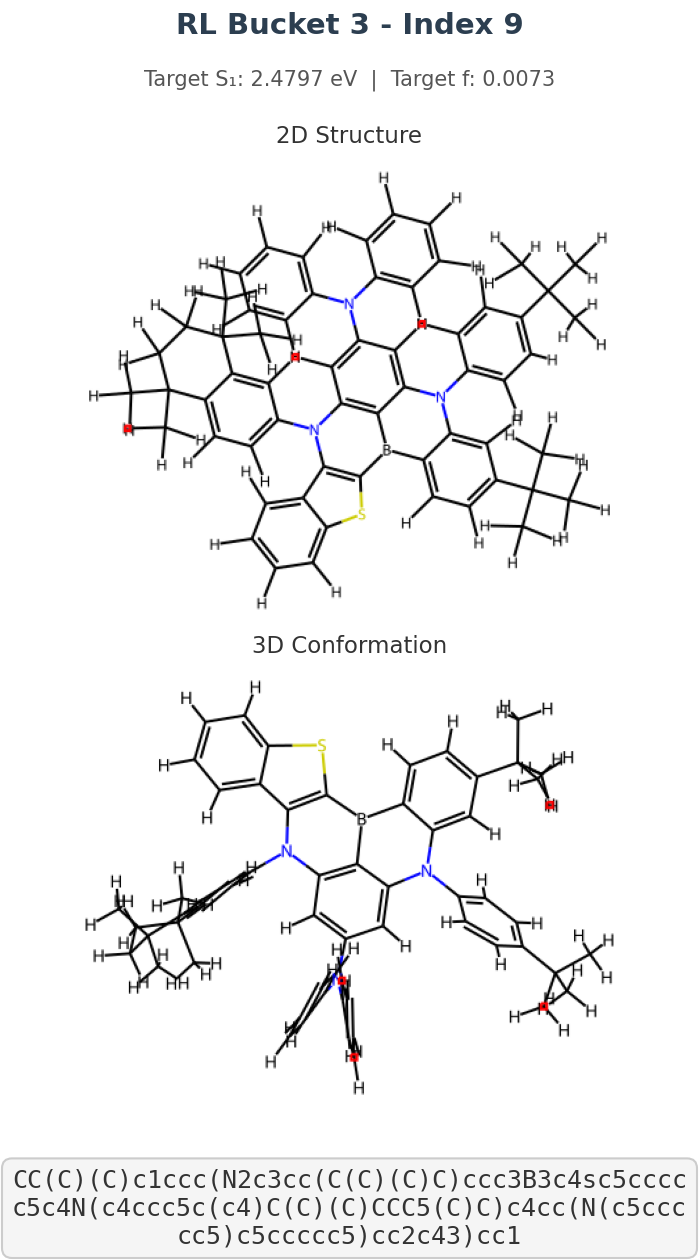}
\end{minipage}
\caption{Generated OLED molecules from Bucket 3 (10).}
\label{fig:bucket3d}
\end{figure}

\begin{figure}[htbp]
\centering
\begin{minipage}{0.32\textwidth}
    \centering
    \includegraphics[width=\textwidth]{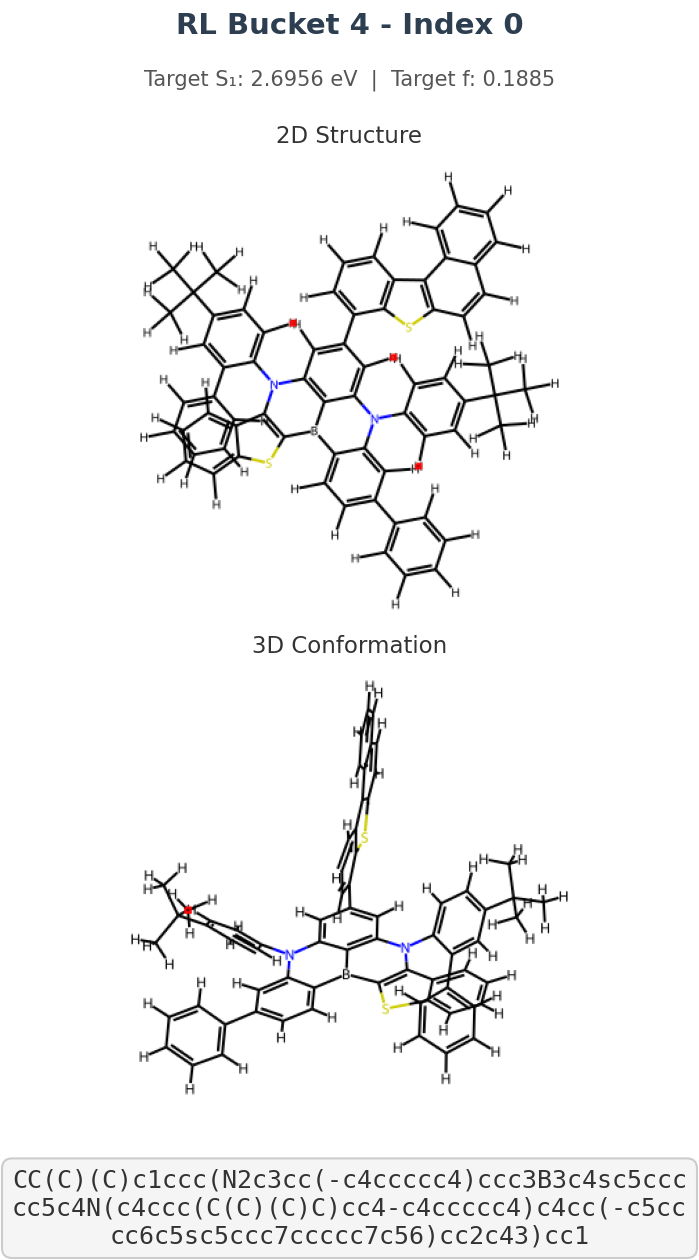}
\end{minipage}
\begin{minipage}{0.32\textwidth}
    \centering
    \includegraphics[width=\textwidth]{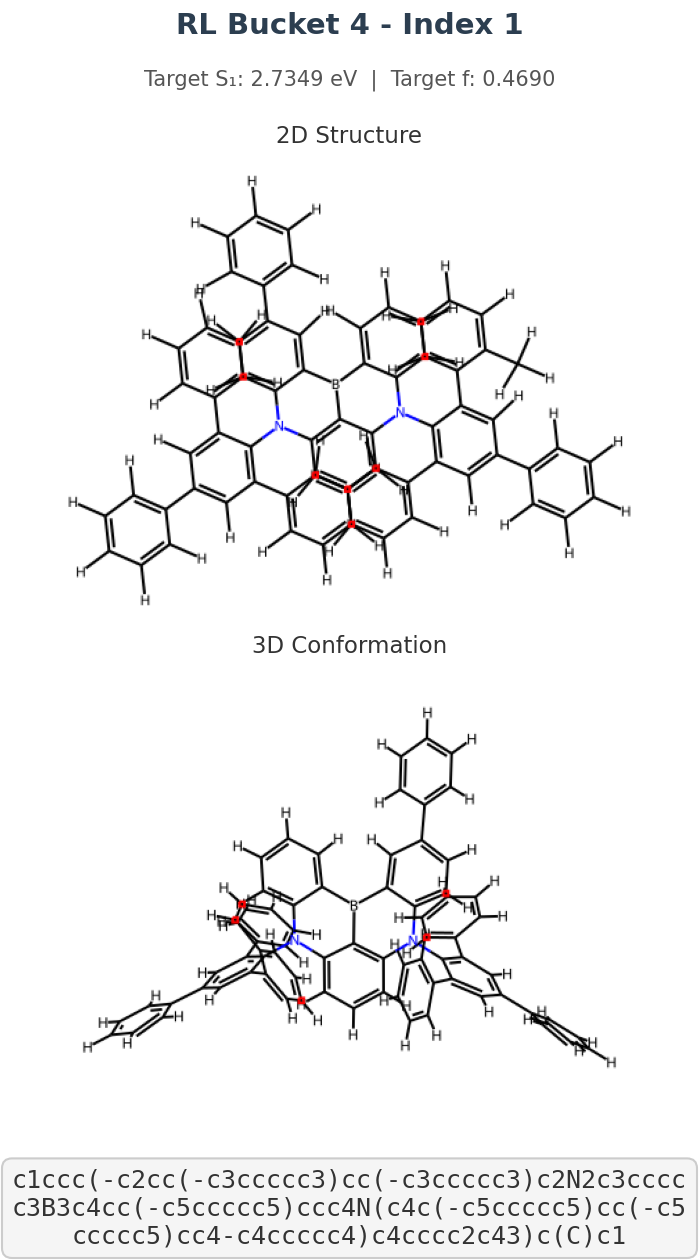}
\end{minipage}
\begin{minipage}{0.32\textwidth}
    \centering
    \includegraphics[width=\textwidth]{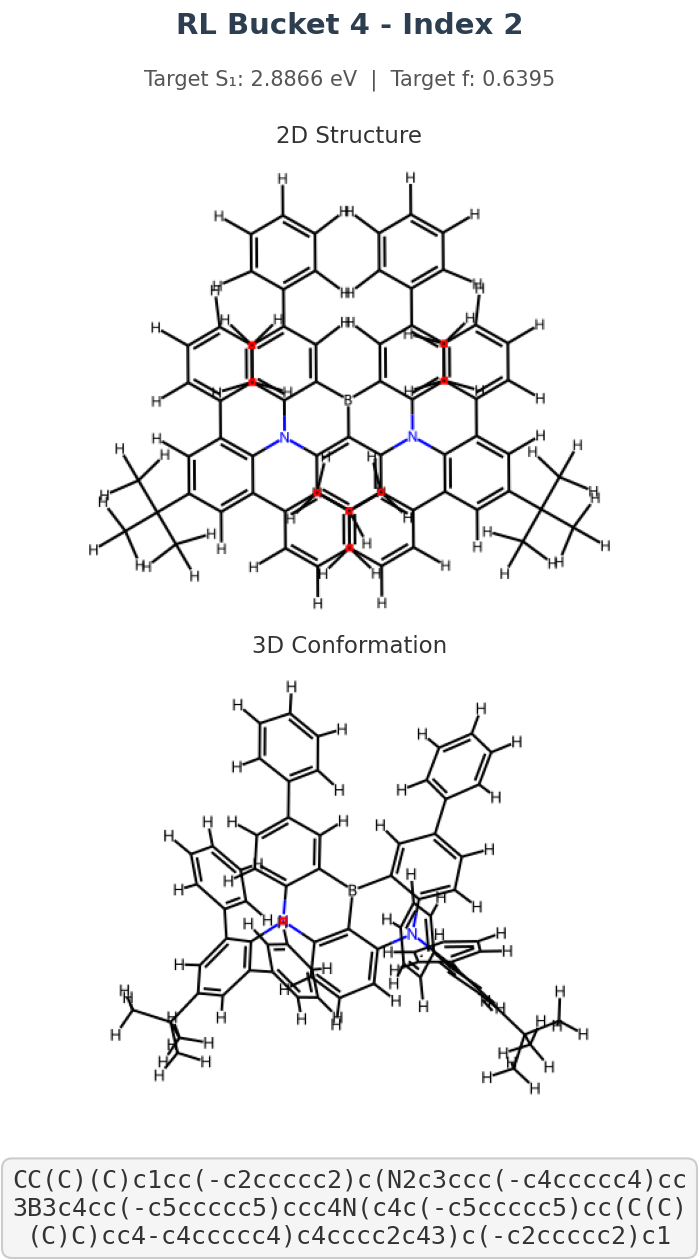}
\end{minipage}
\caption{Generated OLED molecules from Bucket 4 (1-3).}
\label{fig:bucket4a}
\end{figure}

\begin{figure}[htbp]
\centering
\begin{minipage}{0.32\textwidth}
    \centering
    \includegraphics[width=\textwidth]{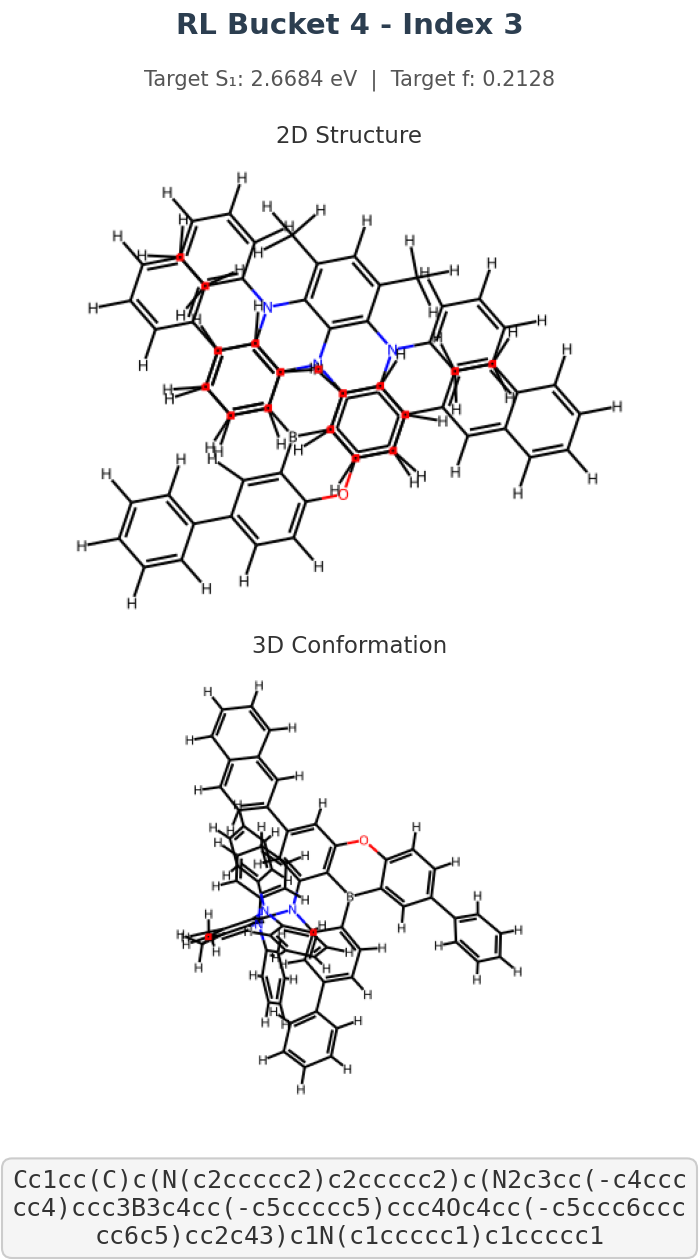}
\end{minipage}
\begin{minipage}{0.32\textwidth}
    \centering
    \includegraphics[width=\textwidth]{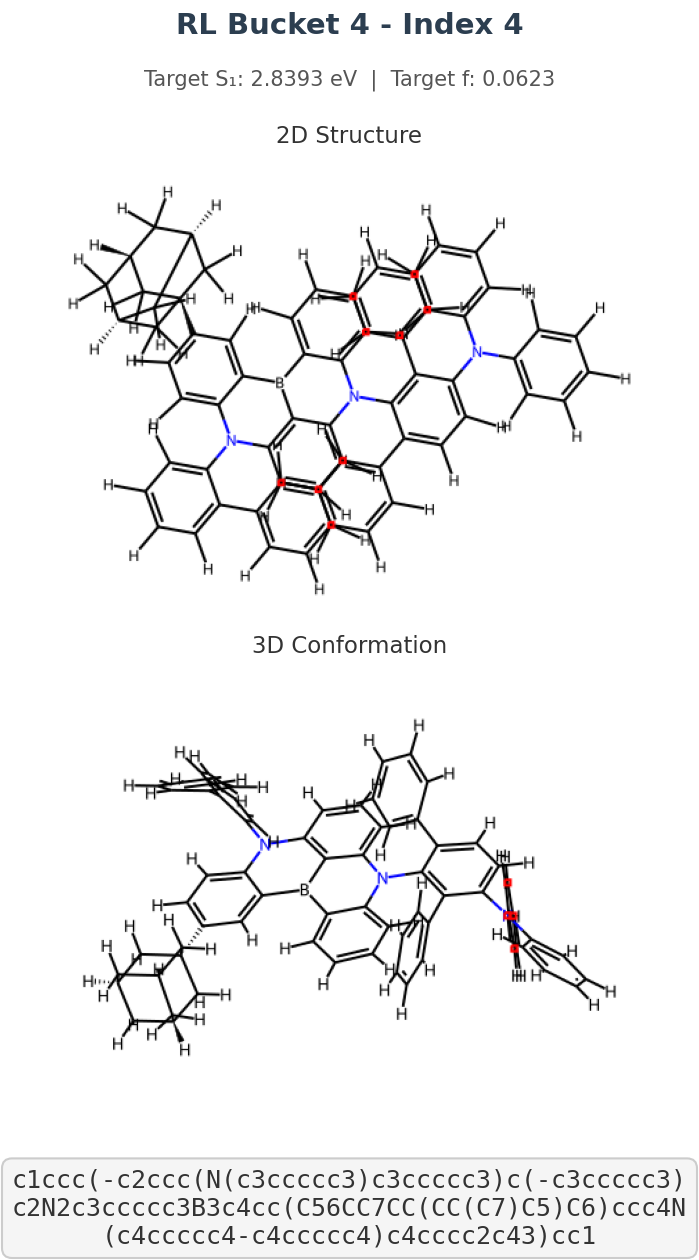}
\end{minipage}
\begin{minipage}{0.32\textwidth}
    \centering
    \includegraphics[width=\textwidth]{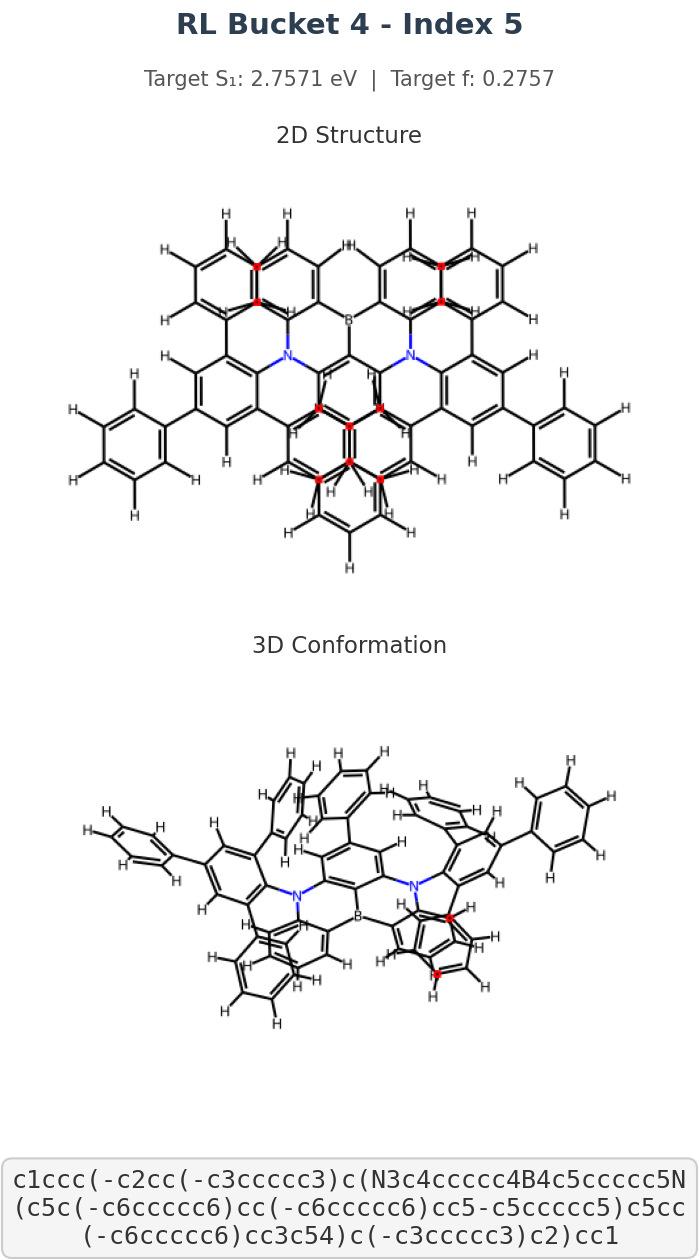}
\end{minipage}
\caption{Generated OLED molecules from Bucket 4 (4-6).}
\label{fig:bucket4b}
\end{figure}

\begin{figure}[htbp]
\centering
\begin{minipage}{0.32\textwidth}
    \centering
    \includegraphics[width=\textwidth]{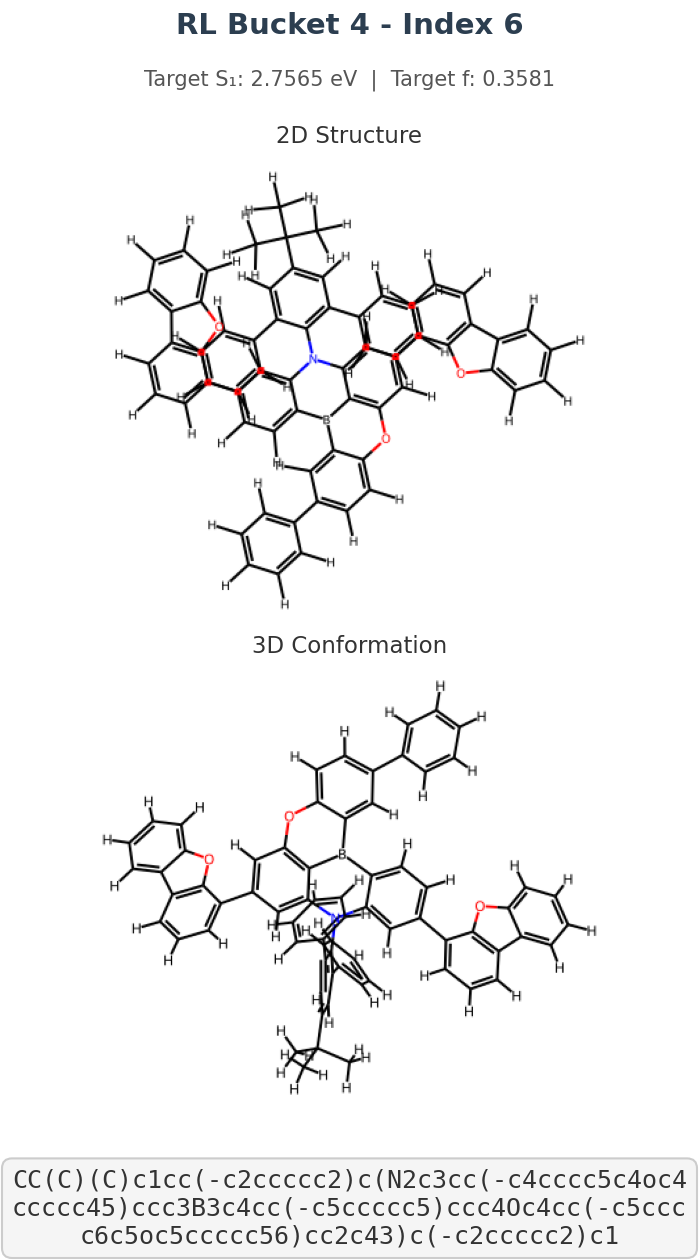}
\end{minipage}
\begin{minipage}{0.32\textwidth}
    \centering
    \includegraphics[width=\textwidth]{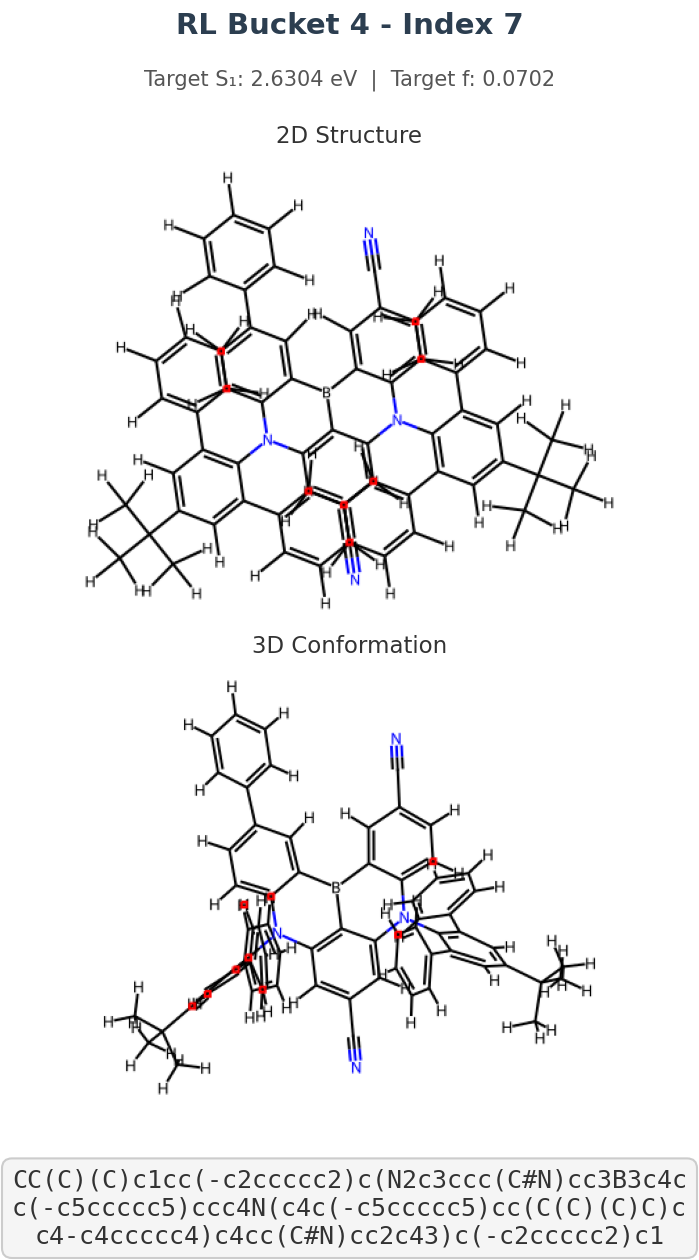}
\end{minipage}
\begin{minipage}{0.32\textwidth}
    \centering
    \includegraphics[width=\textwidth]{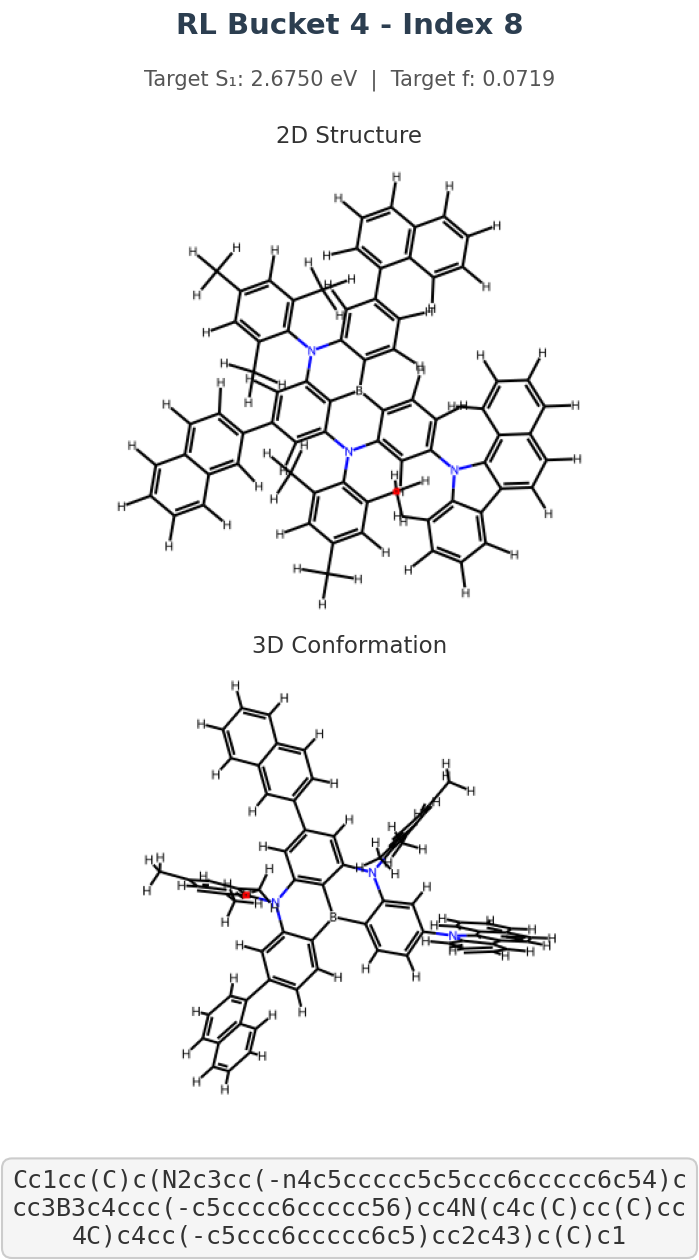}
\end{minipage}
\caption{Generated OLED molecules from Bucket 4 (7-9).}
\label{fig:bucket4c}
\end{figure}

\begin{figure}[htbp]
\centering
\begin{minipage}{0.32\textwidth}
    \centering
    \includegraphics[width=\textwidth]{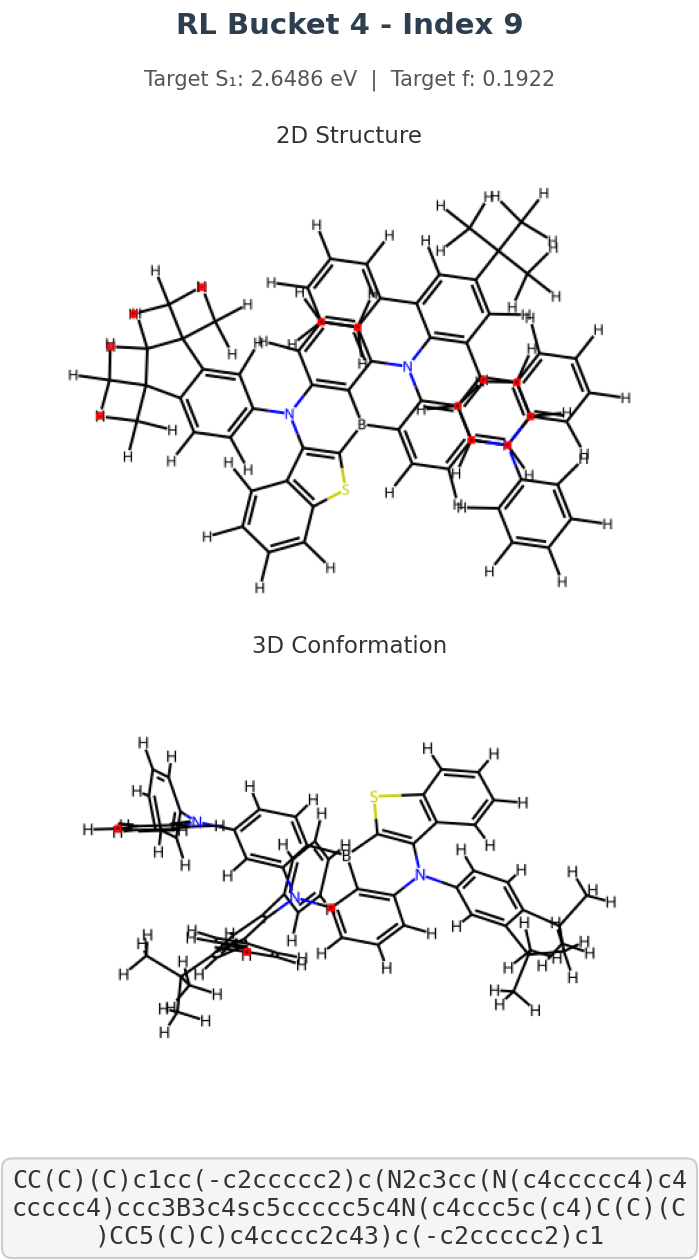}
\end{minipage}
\caption{Generated OLED molecules from Bucket 4 (10).}
\label{fig:bucket4d}
\end{figure}

\begin{figure}[htbp]
\centering
\begin{minipage}{0.32\textwidth}
    \centering
    \includegraphics[width=\textwidth]{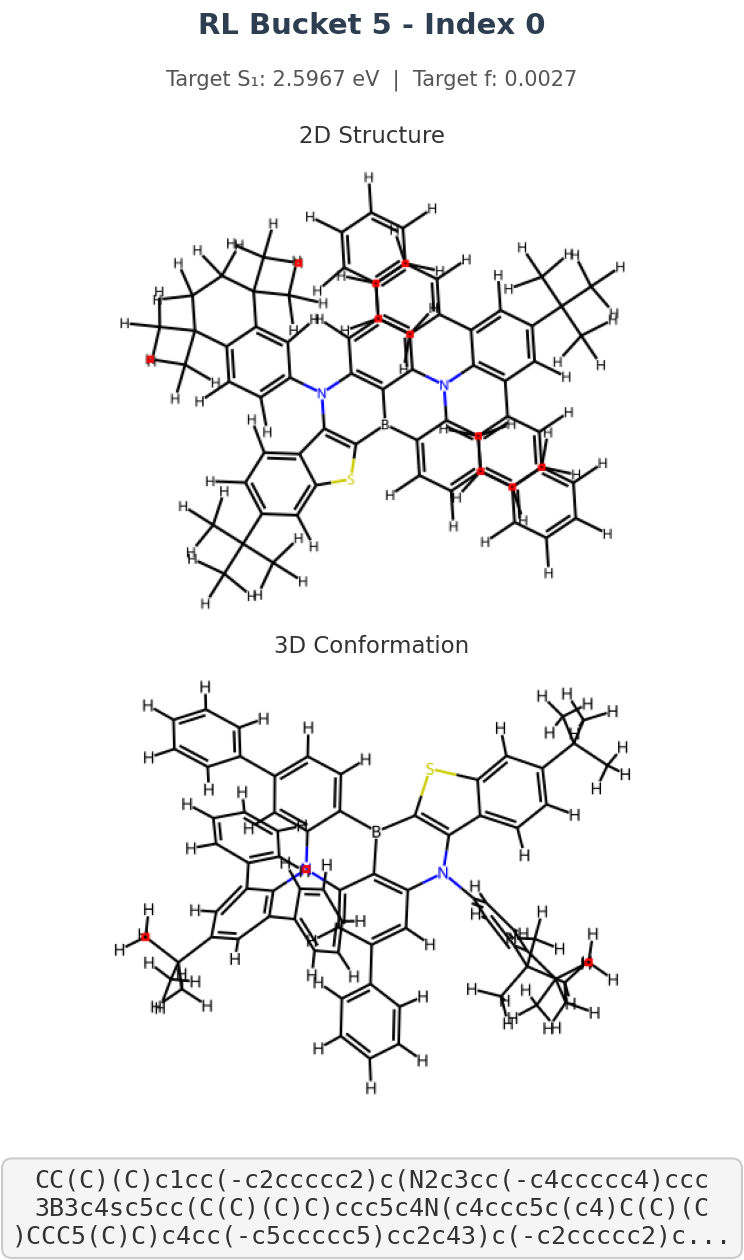}
\end{minipage}
\begin{minipage}{0.32\textwidth}
    \centering
    \includegraphics[width=\textwidth]{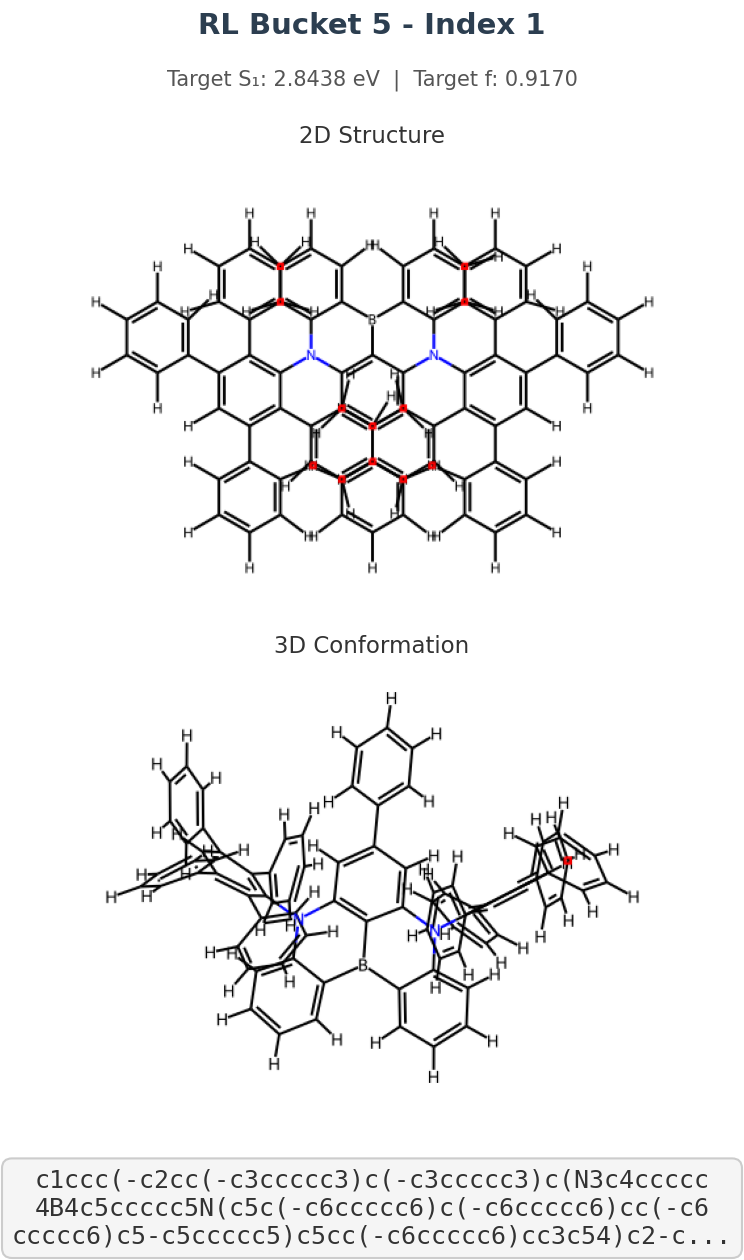}
\end{minipage}
\begin{minipage}{0.32\textwidth}
    \centering
    \includegraphics[width=\textwidth]{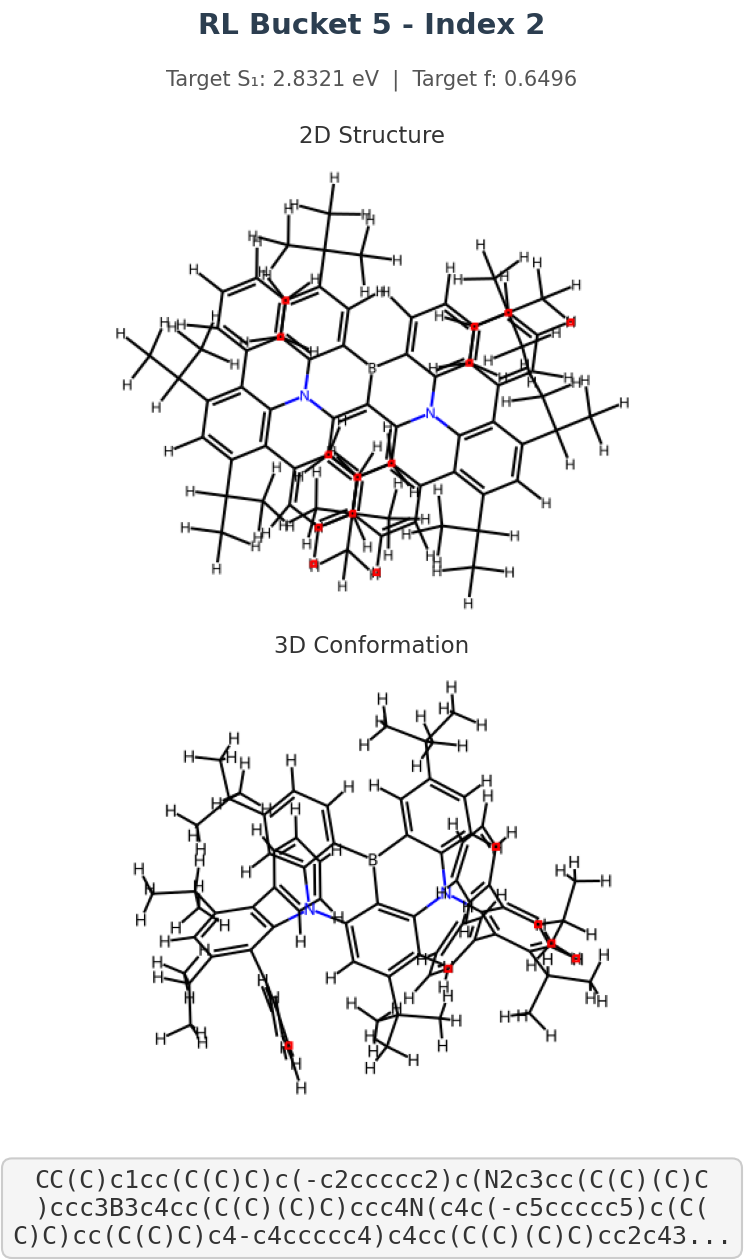}
\end{minipage}
\caption{Generated OLED molecules from Bucket 5 (1-3).}
\label{fig:bucket5a}
\end{figure}

\begin{figure}[htbp]
\centering
\begin{minipage}{0.32\textwidth}
    \centering
    \includegraphics[width=\textwidth]{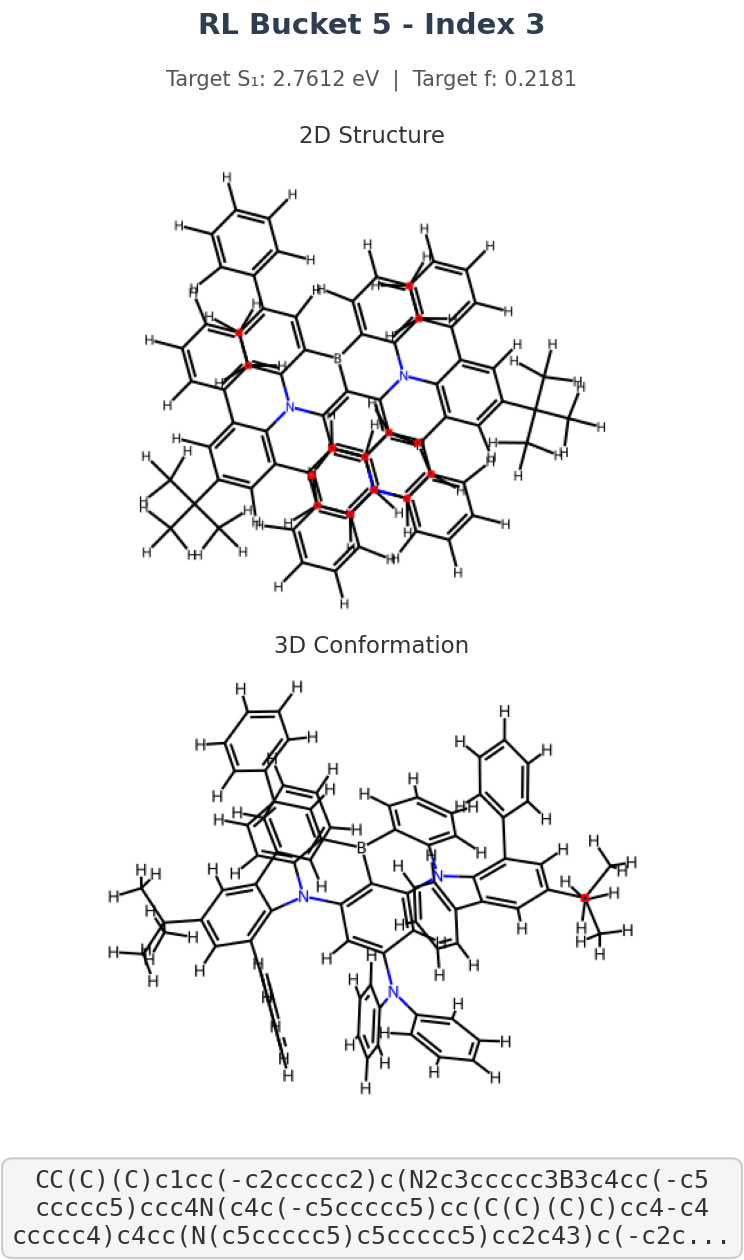}
\end{minipage}
\begin{minipage}{0.32\textwidth}
    \centering
    \includegraphics[width=\textwidth]{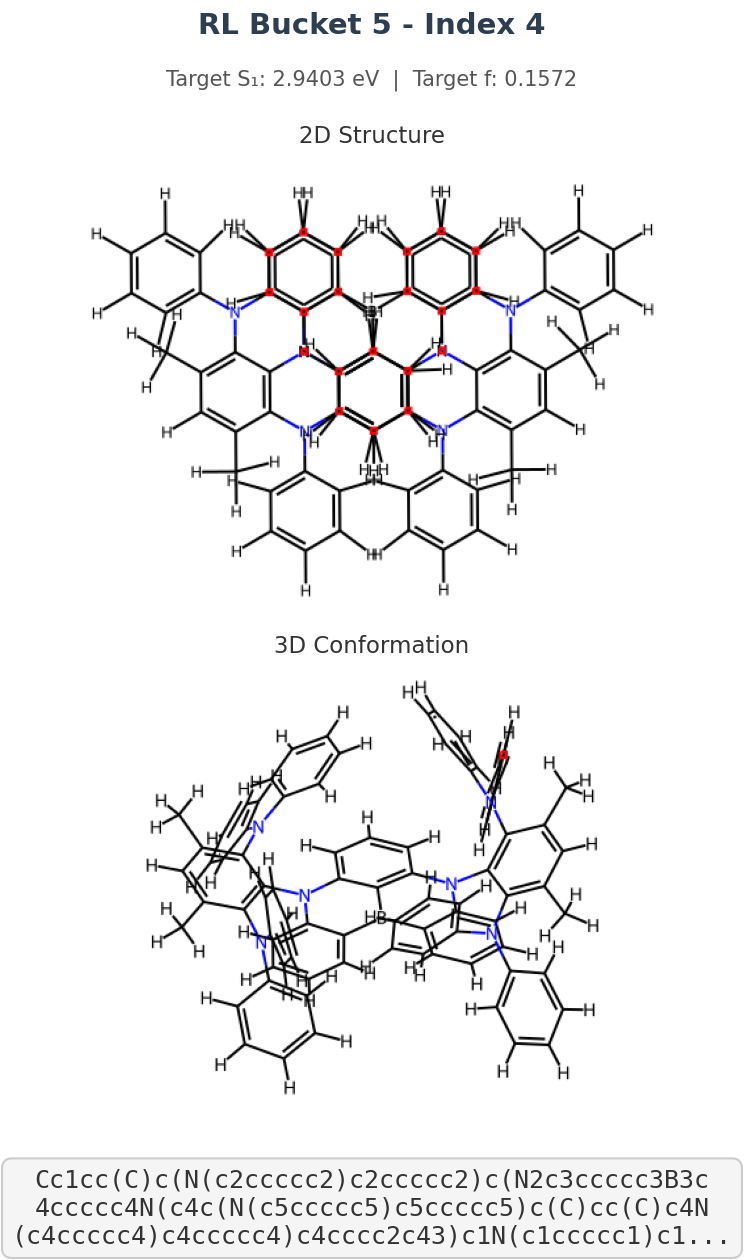}
\end{minipage}
\begin{minipage}{0.32\textwidth}
    \centering
    \includegraphics[width=\textwidth]{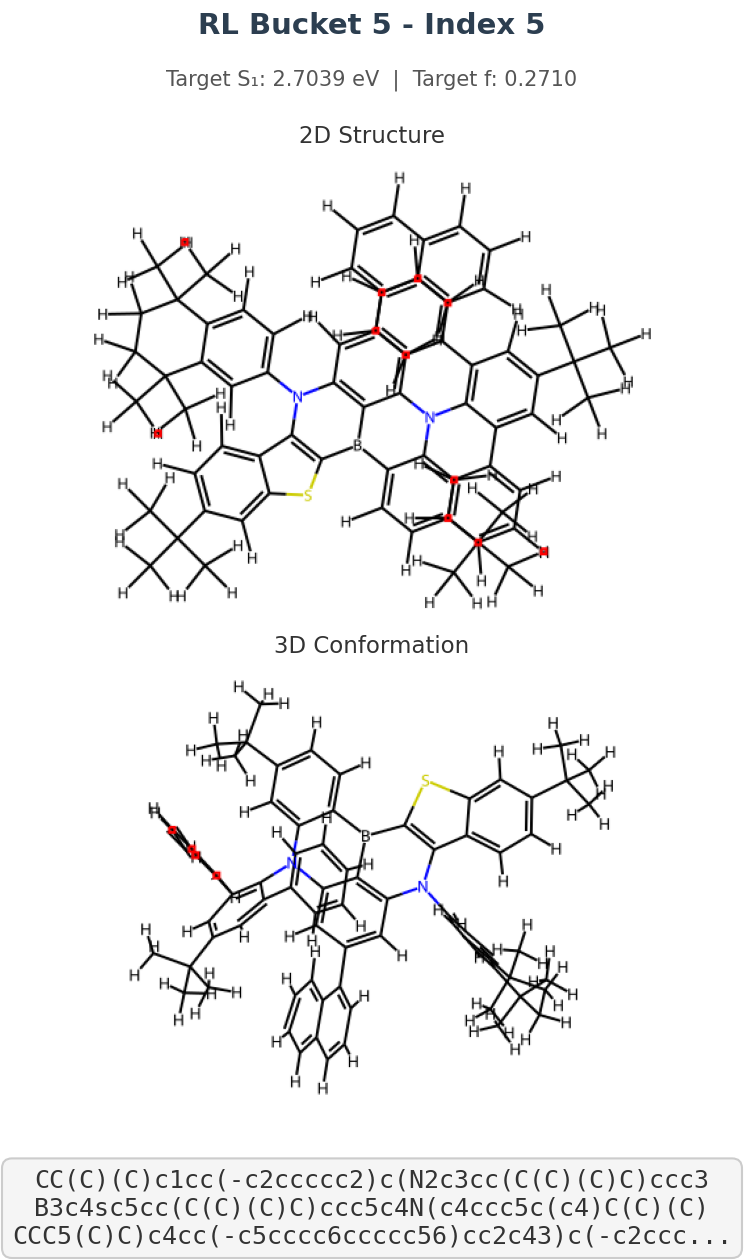}
\end{minipage}
\caption{Generated OLED molecules from Bucket 5 (4-6).}
\label{fig:bucket5b}
\end{figure}

\begin{figure}[htbp]
\centering
\begin{minipage}{0.32\textwidth}
    \centering
    \includegraphics[width=\textwidth]{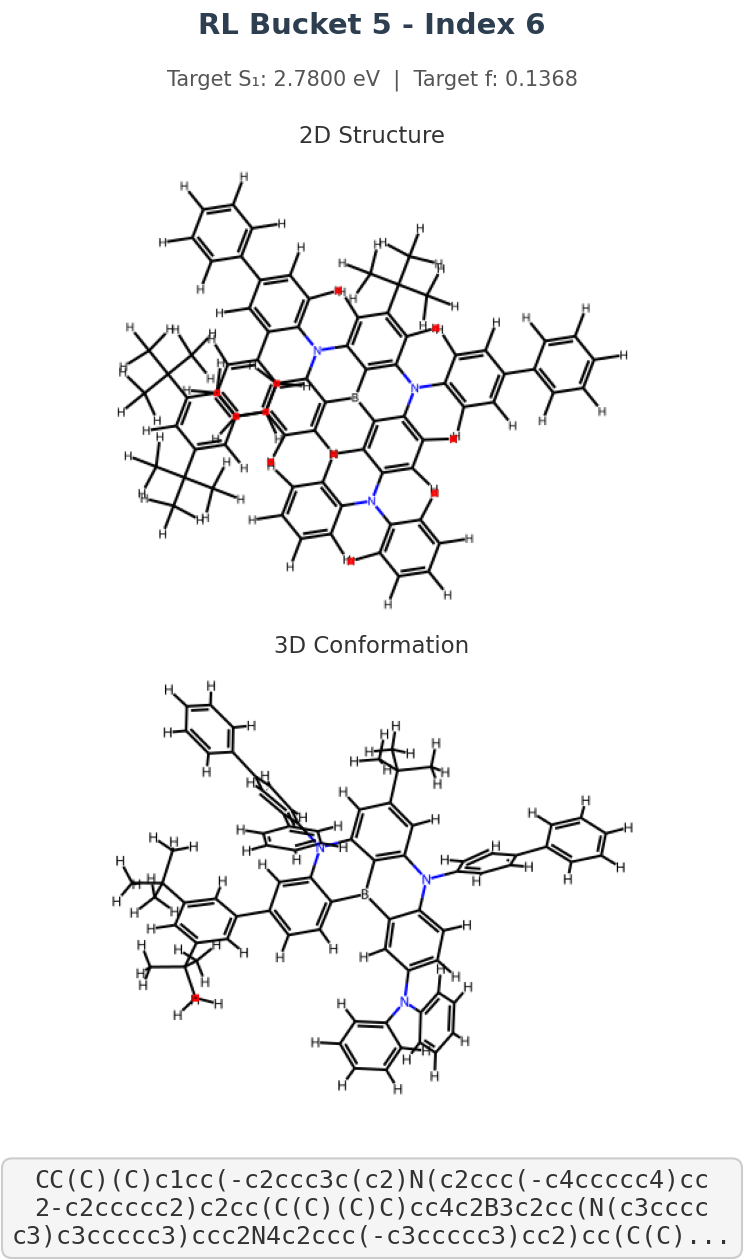}
\end{minipage}
\begin{minipage}{0.32\textwidth}
    \centering
    \includegraphics[width=\textwidth]{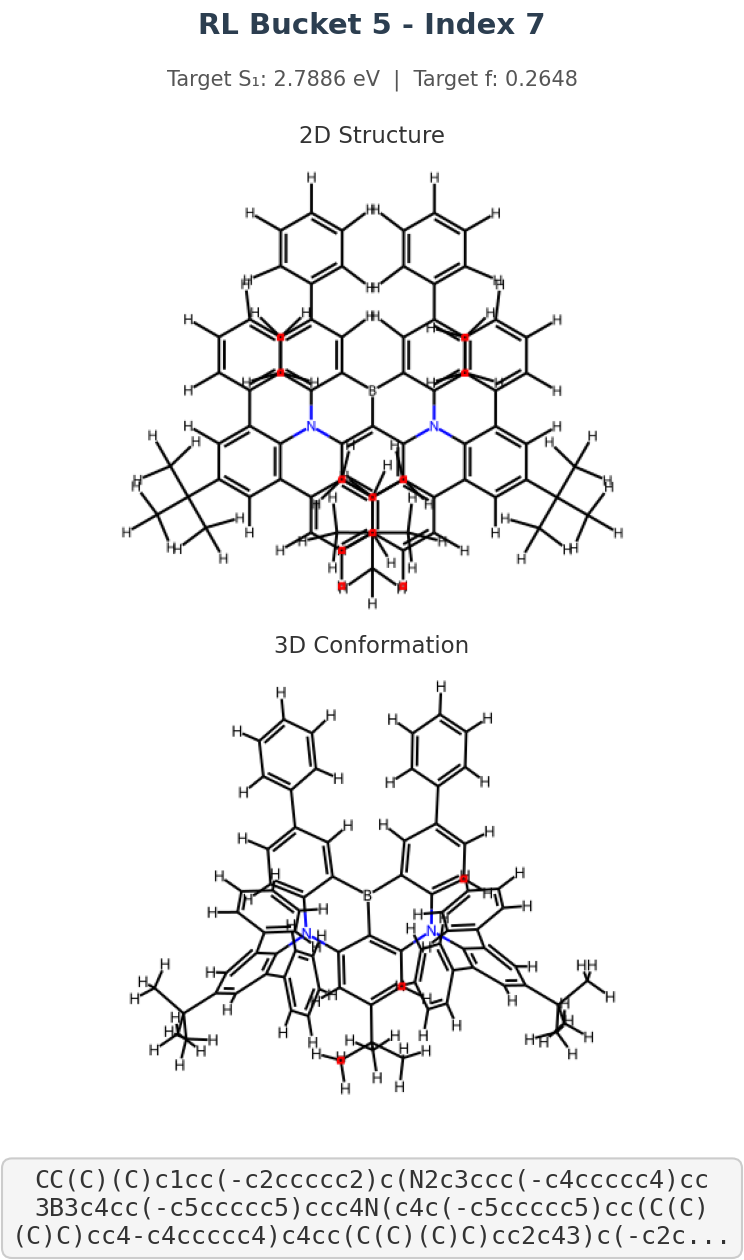}
\end{minipage}
\begin{minipage}{0.32\textwidth}
    \centering
    \includegraphics[width=\textwidth]{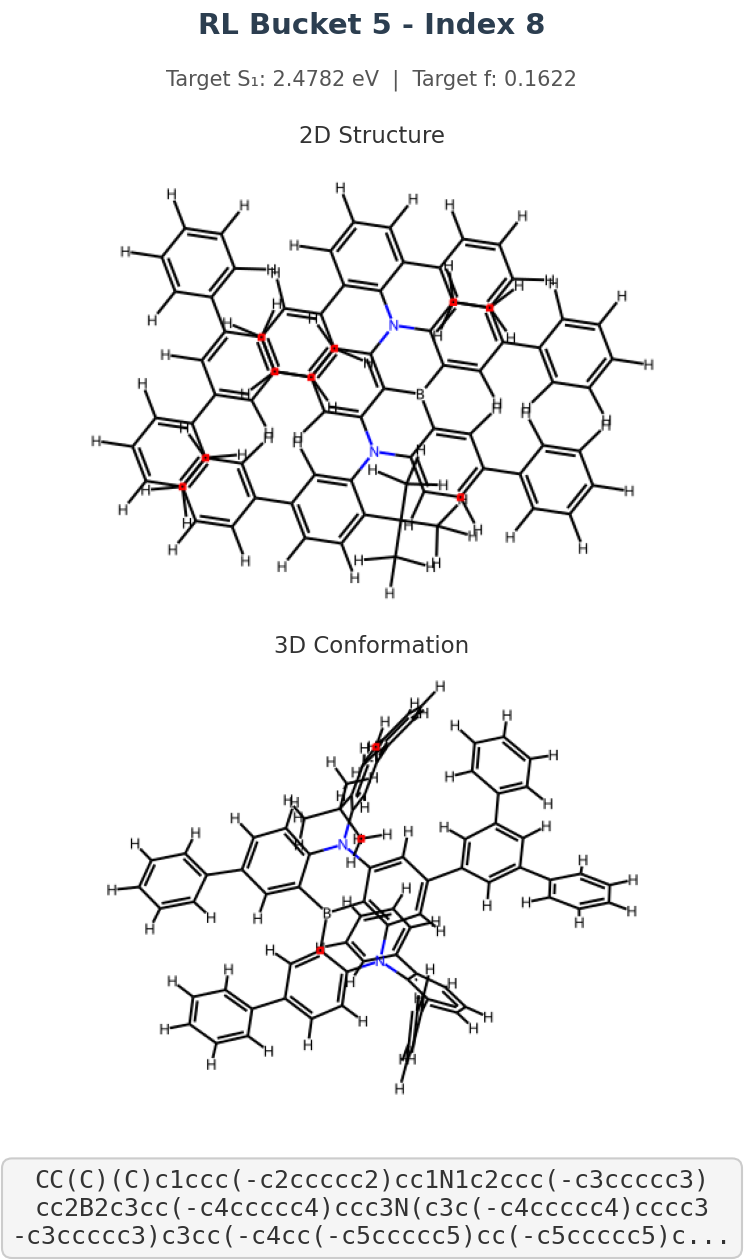}
\end{minipage}
\caption{Generated OLED molecules from Bucket 5 (7-9).}
\label{fig:bucket5c}
\end{figure}

\begin{figure}[htbp]
\centering
\begin{minipage}{0.32\textwidth}
    \centering
    \includegraphics[width=\textwidth]{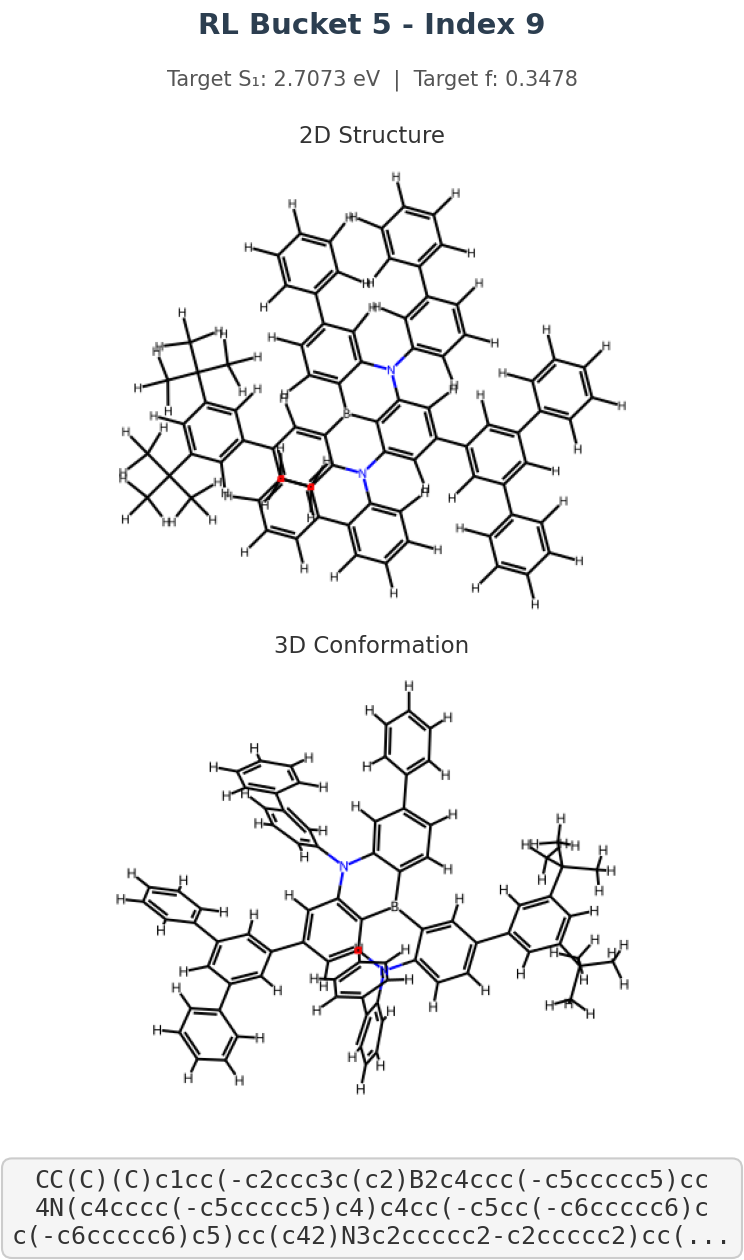}
\end{minipage}
\caption{Generated OLED molecules from Bucket 5 (10).}
\label{fig:bucket5d}
\end{figure}

\section{Pseudocode of Algorithm}
\label{app:pseudocode}

\subsection{BERT}

\begin{algorithm}[H]
\caption{BERT Masked Language Model Pretraining}
\label{alg:bert-mlm}
\begin{algorithmic}[1]
\REQUIRE $\mathcal{D}_{\text{pretrain}}$ (large-scale SMILES dataset), $\rho$ (mask ratio)
\ENSURE $\theta_{\text{bert}}$ (pretrained BERT parameters)
\STATE \textit{Loss: } $\mathcal{L}_{\text{mlm}} = -\sum_{i \in \mathcal{M}} \log P(x_i | \mathbf{x}_{\backslash \mathcal{M}}; \theta_{\text{bert}})$
\STATE Initialize $\theta_{\text{bert}}$ randomly
\FOR{epoch $= 1$ to num\_epochs}
    \FOR{batch $\in$ DataLoader($\mathcal{D}_{\text{pretrain}}$)}
        \STATE $\mathbf{x}_{\text{masked}}, \mathbf{y}, \mathcal{M} \gets \textsc{RandomMask}(\text{batch}, \rho)$
        \STATE $\mathbf{H} \gets \textsc{BERT\_Encoder}(\mathbf{x}_{\text{masked}}; \theta_{\text{bert}})$; $\hat{\mathbf{y}} \gets \textsc{MLM\_Head}(\mathbf{H})$
        \STATE $\mathcal{L}_{\text{mlm}} \gets -\sum_{i \in \mathcal{M}} \log P(\hat{\mathbf{y}}_i = \mathbf{y}_i)$; $\theta_{\text{bert}} \gets \theta_{\text{bert}} - \eta \nabla \mathcal{L}_{\text{mlm}}$
    \ENDFOR
\ENDFOR
\STATE \textbf{return} $\theta_{\text{bert}}$
\end{algorithmic}
\end{algorithm}

\begin{algorithm}[H]
\caption{BERT Property Predictor Fine-tuning}
\label{alg:bert-predictor}
\begin{algorithmic}[1]
\REQUIRE $\mathcal{D}_{\text{labeled}} = \{(\text{smiles}_i, y_i)\}$ (labeled data), $\theta_{\text{bert}}$ (pretrained), $n_{\text{freeze}}$ (frozen layers)
\ENSURE $\theta_{\text{pred}}$ (property predictor parameters)
\STATE \textit{Loss: } $\mathcal{L}_{\text{mse}} = \frac{1}{N} \sum_{i=1}^{N} (\hat{y}_i - y_i)^2$
\STATE $\theta_{\text{pred}} \gets \{\theta_{\text{bert}}, \theta_{\text{head}}\}$ \COMMENT{Freeze first $n_{\text{freeze}}$ layers}
\FOR{epoch $= 1$ to num\_epochs}
    \FOR{$(\text{smiles}, y) \in$ DataLoader($\mathcal{D}_{\text{labeled}}$)}
        \STATE $\mathbf{x} \gets \textsc{Tokenize}(\text{smiles})$; $\mathbf{H} \gets \textsc{BERT\_Encoder}(\mathbf{x}; \theta_{\text{bert}})$
        \STATE $\mathbf{h}_{\text{pool}} \gets \textsc{MeanPooling}(\mathbf{H})$; $\hat{y} \gets \textsc{RegressionHead}(\mathbf{h}_{\text{pool}}; \theta_{\text{head}})$
        \STATE $\mathcal{L}_{\text{mse}} \gets (\hat{y} - y)^2$; $\theta_{\text{pred}} \gets \theta_{\text{pred}} - \eta \nabla \mathcal{L}_{\text{mse}}$
    \ENDFOR
\ENDFOR
\STATE \textbf{return} $\theta_{\text{pred}}$
\end{algorithmic}
\end{algorithm}

\subsection{LLaMA}

\begin{algorithm}[H]
\caption{LLaMA Autoregressive Pretraining}
\label{alg:llama-pretrain}
\begin{algorithmic}[1]
\REQUIRE $\mathcal{D}_{\text{pretrain}}$ (large-scale SMILES dataset)
\ENSURE $\theta_{\text{llama}}$ (pretrained LLaMA parameters)
\STATE \textit{Loss: } $\mathcal{L}_{\text{ntp}} = -\sum_{t=1}^{T} \log P(x_t | x_{<t}; \theta)$
\STATE Initialize $\theta_{\text{llama}}$ randomly
\FOR{epoch $= 1$ to num\_epochs}
    \FOR{batch $\in$ DataLoader($\mathcal{D}_{\text{pretrain}}$)}
        \STATE $\mathbf{x} \gets [\texttt{BOS}] \oplus \textsc{Tokenize}(\text{batch}) \oplus [\texttt{EOS}]$
        \STATE $\mathbf{x}_{\text{in}} \gets \mathbf{x}[:-1]$; $\mathbf{y} \gets \mathbf{x}[1:]$
        \STATE $\hat{\mathbf{y}} \gets \textsc{LLaMA}(\mathbf{x}_{\text{in}}; \theta_{\text{llama}})$
        \STATE $\mathcal{L}_{\text{ntp}} \gets -\sum_{t} \log P(\hat{\mathbf{y}}_t = \mathbf{y}_t)$; $\theta_{\text{llama}} \gets \theta_{\text{llama}} - \eta \nabla \mathcal{L}_{\text{ntp}}$
    \ENDFOR
\ENDFOR
\STATE \textbf{return} $\theta_{\text{llama}}$
\end{algorithmic}
\end{algorithm}

\begin{algorithm}[H]
\caption{LLaMA Conditional Generation Fine-tuning (SFT)}
\label{alg:llama-sft}
\begin{algorithmic}[1]
\REQUIRE $\mathcal{D}_{\text{cond}} = \{(\mathbf{p}_i, \text{smiles}_i)\}$ (conditional data), $\theta_{\text{llama}}$ (pretrained)
\ENSURE $\theta_{\text{sft}}$ (fine-tuned parameters)
\STATE \textit{Loss: } $\mathcal{L}_{\text{sft}} = -\sum_{t=1}^{T} \log P(x_t | \mathbf{c}, x_{<t}; \theta)$
\STATE $\theta_{\text{sft}} \gets \textsc{ExtendVocab}(\theta_{\text{llama}})$ \COMMENT{Add property tokens}
\FOR{epoch $= 1$ to num\_epochs}
    \FOR{$(\mathbf{p}, \text{smiles}) \in$ DataLoader($\mathcal{D}_{\text{cond}}$)}
        \STATE $\mathbf{c} \gets \textsc{EncodeProperties}(\mathbf{p})$
        \STATE $\mathbf{x} \gets \mathbf{c} \oplus [\texttt{BOS}] \oplus \textsc{Tokenize}(\text{smiles}) \oplus [\texttt{EOS}]$
        \STATE $\mathbf{x}_{\text{in}} \gets \mathbf{x}[:-1]$; $\mathbf{y} \gets \mathbf{x}[1:]$
        \STATE $\hat{\mathbf{y}} \gets \textsc{LLaMA}(\mathbf{x}_{\text{in}}; \theta_{\text{sft}})$
        \STATE $\mathcal{L}_{\text{sft}} \gets -\sum_{t} \log P(\hat{\mathbf{y}}_t = \mathbf{y}_t)$; $\theta_{\text{sft}} \gets \theta_{\text{sft}} - \eta \nabla \mathcal{L}_{\text{sft}}$
    \ENDFOR
\ENDFOR
\STATE \textbf{return} $\theta_{\text{sft}}$
\end{algorithmic}
\end{algorithm}

\subsection{GRPO (Group Relative Policy Optimization)}

\begin{algorithm}[H]
\caption{GRPO with BERT Predictor Rewards}
\label{alg:grpo}
\begin{algorithmic}[1]
\REQUIRE $\theta_{\text{sft}}$ (SFT model), $\theta_{\text{pred}}$ (BERT predictors), $\mathcal{D}_{\text{train}}$ (training data), $G$ (group size), $\beta$ (KL weight)
\ENSURE $\theta_{\text{policy}}$ (optimized policy)
\STATE $\theta_{\text{policy}} \gets \theta_{\text{sft}}$, $\theta_{\text{ref}} \gets \theta_{\text{sft}}$ (frozen)
\FOR{step $= 1$ to num\_steps}
    \STATE \textit{// Sampling and Generation}
    \FOR{$i = 1$ to batch\_size}
        \STATE $\mathbf{p}_i \sim \mathcal{D}_{\text{train}}$, $\mathbf{c}_i \gets \textsc{EncodeProperties}(\mathbf{p}_i)$
        \FOR{$g = 1$ to $G$}
            \STATE $s_{i,g} \gets \textsc{Generate}(\mathbf{c}_i; \theta_{\text{policy}})$; $\mathcal{S} \gets \mathcal{S} \cup \{(s_{i,g}, \mathbf{p}_i)\}$
        \ENDFOR
    \ENDFOR
    \STATE \textit{// Reward: } $R(x) = \mathbb{I}_{\text{valid}} [ -\alpha |S_1 - S_1^*| - \gamma |f - f^*|] + (1-\mathbb{I}_{\text{valid}}) \lambda$
    \FOR{$(s, \mathbf{p}) \in \mathcal{S}$}
        \STATE $\tilde{s} \gets \textsc{ExtractSMILES}(s)$
        \STATE $r \gets \lambda_{\text{penalty}}$ \textbf{if} $\neg\textsc{Valid}(\tilde{s})$ \textbf{else} $-\alpha |S_1(\tilde{s}) - S_1^*| - \gamma |f(\tilde{s}) - f^*|$
        \STATE $\mathbf{r}.\text{append}(r)$
    \ENDFOR
    \STATE \textit{// Advantage: } $A_i = (R(o_i) - \mu_g) / (\sigma_g + \epsilon)$
    \STATE $\mathbf{R} \gets \textsc{Reshape}(\mathbf{r}, [B, G])$; $\mu_g, \sigma_g \gets \textsc{Mean}(\mathbf{R}), \textsc{Std}(\mathbf{R})$
    \STATE $\mathbf{A} \gets (\mathbf{R} - \mu_g) / (\sigma_g + \epsilon)$; $\mathbf{a} \gets \textsc{Flatten}(\mathbf{A})$
    \STATE \textit{// Policy Optimization}
    \STATE $\log \pi_{\theta}, \log \pi_{\text{ref}} \gets \textsc{LogProb}(\mathcal{S}; \theta_{\text{policy}}), \textsc{LogProb}(\mathcal{S}; \theta_{\text{ref}})$
    \STATE $\rho_i \gets \exp(\log \pi_{\theta} - \log \pi_{\theta_{\text{old}}})$; $D_{\text{KL}} \gets |\log \pi_{\theta} - \log \pi_{\text{ref}}|$
    \STATE $\mathcal{L} \gets -\frac{1}{G}\sum_i \min(\rho_i A_i, \text{clip}(\rho_i) A_i) + \beta D_{\text{KL}}$
    \STATE \textit{// Parameter Update}
    \STATE $\theta_{\text{policy}} \gets \theta_{\text{policy}} - \eta \nabla \mathcal{L}$
\ENDFOR
\STATE \textbf{return} $\theta_{\text{policy}}$
\end{algorithmic}
\end{algorithm}

\end{document}